\documentclass[11pt]{article}
\usepackage{acl}  
\usepackage{times}
\usepackage{latexsym}
\usepackage[T1]{fontenc}
\usepackage[utf8]{inputenc}
\usepackage{microtype}
\usepackage{graphicx}
\usepackage{booktabs}
\usepackage{multirow}
\usepackage{amsmath,amssymb}
\usepackage{enumitem}
\usepackage{capt-of}
\usepackage[ruled,noend]{algorithm2e}
\usepackage{float}
\usepackage{tikz}
\usetikzlibrary{positioning,arrows.meta}

\graphicspath{{figures/}}

\title{Temporal Multi-Signal Fusion for Token-Level Hallucination Detection}

\author{Igor Itkin \\
  Independent Researcher \\
  \texttt{ig.itkin@gmail.com} \\
  ORCID: 0009-0004-9513-8463 \\
  Preprint, June 2026}

\begin{document}
\maketitle

\begin{abstract}
Token-level hallucination detectors score each token independently from a single signal, and fail exactly when the generating model is confidently wrong.
This paper instead treats hallucination as a temporally extended span and detects it by sequence labeling: each token is scored from a 33-dimensional feature stream that fuses text statistics, Natural Language Inference (NLI) entailment, and language model surprisal, with no access to model internals.
A Bidirectional Gated Recurrent Unit (BiGRU) over these features reaches an AUC of $0.840$ on RAGTruth (10 seeds), an 11-point gain over an independent logistic-regression baseline ($p = 0.002$, Wilcoxon signed-rank).
A controlled decomposition attributes most of the gain to temporal order rather than model capacity: evidence propagates from confident positions to ambiguous neighbors within a span.
The same $0.845$ ceiling recurs across recurrent, state-space (Mamba), and attention architectures, locating the bottleneck in the feature set rather than the model.
Because it reads only the generated text and external signals, the detector works on closed-source models, and it keeps working on text produced by language models it never saw during training, losing under $4\%$ AUC.

\end{abstract}

\section{Introduction}
\label{sec:introduction}

Large language models generate text that is fluent, contextually appropriate, and sometimes entirely fabricated~\cite{ji2023survey,huang2023survey}.
In retrieval-augmented generation systems, a single hallucinated claim can propagate through downstream decisions in healthcare, legal reasoning, or financial analysis.
The core difficulty is that these models fail silently: there is no intrinsic signal that anything has gone wrong.

A substantial body of work addresses this problem from different angles.
Internal probes train classifiers on hidden representations to separate hallucinated from grounded tokens within the model's own representation space~\cite{kossen2024seps,chen2024inside}.
Uncertainty estimators measure entropy or semantic divergence across sampled generations, using the model's own confidence as a proxy for correctness~\cite{farquhar2024semanticentropy}.
Natural Language Inference (NLI)-based classifiers compare generated claims against source documents using external entailment models~\cite{lettucedetect2025,halugate2025}.

Each of these approaches makes progress, yet each has a characteristic failure mode.
\citet{zhang2025trustmewrong} show that uncertainty-based detectors fail on high-confidence hallucinations, precisely the cases that matter most.
\citet{cheng2025llmsdonotknow} demonstrate that internal probes become indistinguishable when the model has partial knowledge of the subject.
\citet{dubanowska2025oodgeneralization} find that current detectors do not generalize out-of-distribution.
These failures are not coincidental: all existing methods share a common design choice that limits their effectiveness.
Every approach examines each token in isolation, computing a score at one position without reference to how the signal at neighboring positions is evolving.

We argue that this per-token independence assumption discards a fundamental regularity: hallucination is not a point event.
When an autoregressive model departs from factual grounding, the erroneous tokens become part of the context for subsequent generation, biasing the model toward continuation of the hallucinated narrative.
The result is characteristic runs of hallucinated tokens whose statistical signatures evolve jointly over time.
In RAGTruth~\cite{niu2024ragtruth}, we confirm a median span of 5~tokens and a persistence probability $P(H_t \mid H_{t-1}) = 0.902$ versus an onset probability $P(H_t \mid F_{t-1}) = 0.006$---a 150:1 ratio.
An information-theoretic analysis shows that 76\% of label entropy is predictable from one step of temporal context.

We exploit this structure by casting token-level detection as sequence labeling.
For each token, we construct a 33-dimensional feature vector fusing three signal families: text statistics (context overlap, novelty dynamics), NLI entailment from DeBERTa~\cite{he2021deberta}, and language model surprisal from TinyLlama~\cite{zhang2024tinyllama}.
Each signal is enriched with temporal derivatives (running means, first-order differences, windowed extremes).
A BiGRU sequence labeler then learns the joint temporal dynamics across positions.

We organize our investigation around three research questions:

\begin{description}[nosep,leftmargin=1em]
\item[RQ1] Does modeling temporal dependencies between token positions improve hallucination detection compared to independent per-token classification?
\item[RQ2] Does fusing multiple signal types (text, NLI, language model) outperform any single signal family?
\item[RQ3] Do the learned temporal patterns generalize across source models and datasets?
\end{description}

Our main contributions are as follows:
\begin{enumerate}[nosep,leftmargin=1.5em]
\item We frame hallucination detection as sequence labeling with temporal dependencies and show that BiGRU achieves 0.840 AUC, outperforming independent classifiers by 11 points. A controlled decomposition isolates the contribution of temporal order (44\%), sequence aggregation (24\%), and nonlinear capacity (32\%).
\item We introduce multi-signal temporal fusion: 33 features from three complementary families, enriched with temporal statistics. No single signal exceeds 0.641 AUC alone; their fusion under temporal modeling reaches 0.840.
\item We provide a generalization study across six source models (3.8\% relative degradation) and two datasets. Annotation density matters more than dataset size for cross-domain transfer.
\item We show that Conditional Random Field models scored with softmax are severely miscalibrated; forward-backward marginals recover up to +17.9 AUC points.
\end{enumerate}

\section{Related Work}
\label{sec:related_work}

Hallucination detection methods fall into four families, distinguished by the signal they use and the access they require.
We survey each family, focusing on methods evaluated on our primary benchmark RAGTruth~\cite{niu2024ragtruth}, and identify the structural limitation common to all of them.

\paragraph{Internal probes.}
Linear probes on hidden states~\cite{kossen2024seps,chen2024inside}, sparse autoencoder features~\cite{hallusae2026,raglens2025,ferrando2024entityknowledge}, and jointly trained detection heads~\cite{singledirection2025} can separate hallucinated tokens in the model's representation space.
These methods require access to the generating model's internals, which is unavailable for closed-source APIs.
Even in white-box settings, \citet{cheng2025llmsdonotknow} show that internal representations become indistinguishable under partial knowledge, and \citet{roy2026detection} report a robust asymmetry whereby activation probes can flag hallucination yet cannot steer the model to correct it, with the signal concentrated early and emerging only at larger scales.
Recent work selects the most informative intermediate layer for such probes~\cite{wang2026fepoid}; like most probing methods it detects at the response level, whereas we operate at the token level, where a probe must localize each token rather than judge a whole response and hidden-state probes score correspondingly lower.
A related line treats the generator as a dynamical system and probes its stability, for example Lyapunov probes at knowledge-transition boundaries~\cite{luan2026lyapunov}, but still requires model internals.

\paragraph{Uncertainty and logit-based methods.}
Semantic entropy~\cite{farquhar2024semanticentropy} measures distributional divergence across sampled generations.
At the token level, entropy production rate~\cite{entropyproduction2025} tracks uncertainty evolution, while HIDE~\cite{hide2025} flags hallucinations from the statistical decoupling between a model's input-context and output representations in a single forward pass.
Closest in spirit to our temporal view, HALT~\cite{shapiro2026halt} treats a generation's log-probabilities as a time series and reads them with a recurrent network, but it uses a single signal family (log-probs) and emits one response-level verdict, whereas we fuse text, NLI, and log-probability signals and localize hallucination per token.
\citet{snel2025firsttokens} show on RAGTruth that the first token in a hallucination span is detectable via entropy (AUC near 0.8), while subsequent tokens in the same span are near chance.
This finding is directly relevant to our work: the temporal model can propagate onset evidence forward to maintain detection confidence on continuation tokens.
Mechanistic work supports the forward direction: \citet{akarlar2026trajectory} show by activation patching that a hallucinated state corrupts the continuation far more readily (87.5\%) than the reverse repair restores it (33.3\%), an asymmetry consistent with our forward temporal modeling.
LM-Polygraph~\cite{fadeeva2024lmpolygraph} provides a unified framework for comparing such pointwise uncertainty estimates.

\paragraph{Token-level classification and NLI.}
The closest existing work to ours frames detection as token or span classification.
LettuceDetect~\cite{lettucedetect2025} fine-tunes a 395M-parameter ModernBERT encoder and achieves span F1 of 58.9\% on RAGTruth.
HaluGate~\cite{halugate2025} combines sentinel classification with NLI-based explanation.
\citet{ogasa2025rl4hs} train 7--14B generative models with reinforcement learning, reaching span F1 of 58.3\%.
Concurrent work scales token-level detection through synthetic data engines and importance weighting~\cite{min2026tokenhd}, and \citet{obeso2025realtime} detect hallucinated entities in real time during long-form generation.
These systems classify each span or token independently, without modeling dependencies between neighboring positions.

\paragraph{Activation-based methods.}
ReDeEP~\cite{redeep2024} extracts proxy model activations and achieves 0.733 AUC on RAGTruth.
RagtStacking~\cite{ragtstacking2025} extends this with calibrated stacking, reaching 0.836 AUC (Gemma-2-9B), and shows that AUC spans only 2.3 points across an eighteen-fold model size difference.
Lookback Lens~\cite{chuang2024lookback} computes per-token attention ratios from the generator's self-attention maps.
These methods require running an open-weight proxy model and extracting its activations.

All four families share a structural limitation: each token is scored in isolation.
A probe, an entropy score, or an NLI label is computed per position without reference to how the signal at neighboring positions is evolving.
Our approach differs in two respects: we fuse multiple signal types and we model their joint temporal evolution across the sequence.
Table~\ref{tab:positioning} summarizes the positioning.

\begin{table}[t]
\centering
\caption{Positioning on RAGTruth. $^\dagger$Span F1. $^\ddagger$10 seeds. $^\S$0.73 = the published Lookback Lens (LogReg) classifier; 0.838 = the same attention features under our BiGRU sequence model. Both use a Qwen3-14B proxy on the open-LLM subset; see \S\ref{sec:lookback_results} for the full fair comparison.}
\label{tab:positioning}
\small
\begin{tabular}{@{}lccc@{}}
\toprule
\textbf{Method} & \textbf{Access} & \textbf{Metric} & \textbf{Score} \\
\midrule
ReDeEP & Activations & AUC & 0.733 \\
RagtStacking & Activations & AUC & 0.836 \\
Lookback Lens$^\S$ & Attention & AUC & 0.73 / 0.838 \\
LettuceDetect & Black-box & Span F1$^\dagger$ & 0.589 \\
RL4HS & Black-box & Span F1$^\dagger$ & 0.583 \\
HaluGate & Black-box & Token F1 & 0.590 \\
\midrule
\textbf{Ours (BiGRU)} & \textbf{Black-box} & \textbf{AUC}$^\ddagger$ & $\mathbf{0.840 \pm .007}$ \\
\bottomrule
\end{tabular}
\end{table}

We select these baselines because they represent the state of the art across access levels and report results on RAGTruth, which allows direct comparison.
While recurrent models over a single signal exist at the response level~\cite{shapiro2026halt}, our work is, to our knowledge, the first to model token-level temporal dependencies for per-token localization and the first to learn the fusion of multiple complementary signal types rather than relying on a single family.

\section{Method}
\label{sec:method}

\begin{figure}[t]
\centering
\begin{tikzpicture}[
  font=\footnotesize,
  box/.style={draw, rounded corners=1.5pt, align=center, inner sep=3.5pt},
  sig/.style={draw, align=center, inner sep=2.5pt, minimum height=8mm},
  arr/.style={-{Stealth[length=2mm]}, semithick}
]
\node[box, fill=gray!10, text width=0.88\linewidth] (input)
  {\textbf{Input:} source context $C$ \,+\, response $w_1, \dots, w_T$};

\node[sig, fill=blue!8, below=4.5mm of input.south west, anchor=north west, text width=0.26\linewidth] (text)
  {\textbf{Text}\\ overlap, novelty\\ 20 dim};
\node[sig, fill=green!8, below=4.5mm of input.south, anchor=north, text width=0.26\linewidth] (nli)
  {\textbf{NLI}\\ DeBERTa\\ 7 dim};
\node[sig, fill=orange!12, below=4.5mm of input.south east, anchor=north east, text width=0.26\linewidth] (lm)
  {\textbf{LM}\\ TinyLlama\\ 6 dim};

\node[box, fill=gray!10, below=4.5mm of nli, text width=0.88\linewidth] (feat)
  {Feature matrix $\mathbf{X} \in \mathbb{R}^{T \times 33}$\\ with running means, deltas, windowed extremes};

\node[box, fill=violet!10, below=4.5mm of feat, text width=0.88\linewidth] (gru)
  {\textbf{BiGRU} (2 layers, $h{=}64$) $\to$ MLP head};

\node[box, fill=gray!10, below=4.5mm of gru, text width=0.88\linewidth] (out)
  {Per-token probabilities $\hat{y}_t = P(\text{hallucinated} \mid \mathbf{X})$\\ trained with class-weighted BCE (Eq.~\ref{eq:loss})};

\draw[arr] (input.south -| text) -- (text.north);
\draw[arr] (input.south) -- (nli.north);
\draw[arr] (input.south -| lm) -- (lm.north);
\draw[arr] (text.south) -- (feat.north -| text);
\draw[arr] (nli.south) -- (feat.north);
\draw[arr] (lm.south) -- (feat.north -| lm);
\draw[arr] (feat.south) -- (gru.north);
\draw[arr] (gru.south) -- (out.north);
\end{tikzpicture}
\caption{Method overview. Three external signal families are extracted per token (no access to the generating LLM is required), enriched with temporal statistics, and labeled jointly by a bidirectional sequence model.}
\label{fig:pipeline}
\end{figure}
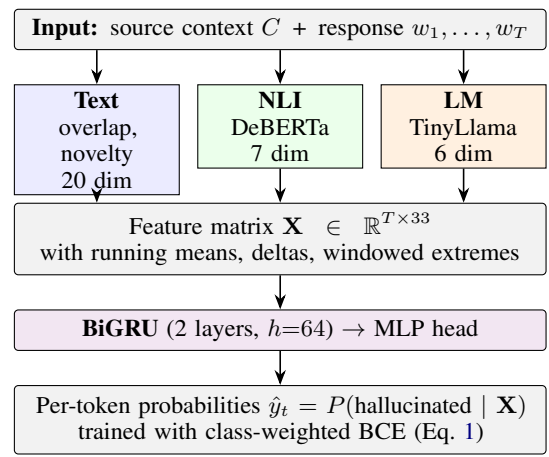

Figure~\ref{fig:pipeline} provides an overview of the approach: per-token features are extracted from three external signals and fed to a bidirectional sequence labeler.

\subsection{Problem Statement}
\label{sec:problem_statement}

Given an LLM-generated response $\mathbf{w} = (w_1, \dots, w_T)$ produced in the context of a source document $C$, we assign each token a binary label $y_t \in \{0, 1\}$ indicating whether $w_t$ is hallucinated.
We construct per-token features $\mathbf{x}_t \in \mathbb{R}^{33}$ (described in Section~\ref{sec:features}) and train a sequence labeler $f_\theta$ that maps the full feature matrix $\mathbf{X} = (\mathbf{x}_1, \dots, \mathbf{x}_T)$ to predicted probabilities $\hat{y}_t = f_\theta(\mathbf{X})_t$.
The model is trained by minimizing binary cross-entropy with class-balanced weighting:
\begin{equation}
\mathcal{L} = -\frac{1}{T} \sum_{t=1}^{T} \left[ \alpha \, y_t \log \hat{y}_t + (1 - y_t) \log(1 - \hat{y}_t) \right]
\label{eq:loss}
\end{equation}
where $\alpha = n_{\text{neg}} / n_{\text{pos}}$ compensates for class imbalance ($\sim$5.6\% positive tokens).
We evaluate with token-level AUC-ROC (ranking quality, threshold-independent) and token-level F1 at threshold 0.5.

\subsection{Feature Extraction}
\label{sec:features}

For each token $w_t$ we extract 33 features from three complementary signal families (Table~\ref{tab:features}).
All features are computed from the generated text and external models only; no access to the generating LLM's internals is required.

\begin{table}[t]
\centering
\small
\begin{tabular}{@{}llc@{}}
\toprule
\textbf{Group} & \textbf{Signals} & \textbf{Dim} \\
\midrule
Text & Surface, overlap, running statistics & 20 \\
NLI  & Entailment scores + temporal derivatives & 7 \\
LM   & Surprisal, rank + fallback handling & 6 \\
\midrule
& \textbf{Total} & \textbf{33} \\
\bottomrule
\end{tabular}
\caption{Per-token feature groups.}
\label{tab:features}
\end{table}

\paragraph{Text features (20 dimensions).}
Surface indicators include word length, numeric and capitalization flags, and absolute and relative position.
Context overlap (a binary indicator $\mathbf{1}[w_t \in C]$) records whether the token appears in the source, extended to bigram and trigram overlap.
Running statistics capture sequential dynamics: cumulative overlap ratio, novelty rate, consecutive novel token count, windowed averages (5, 10, 20 tokens), and first- and second-order differences of the novelty ratio.

\paragraph{NLI features (7 dimensions).}
We compute sentence-level entailment, contradiction, and neutral probabilities using DeBERTa-v3-large~\cite{he2023debertav3} fine-tuned on Multi-Genre NLI~\cite{williams2018mnli}, with contexts truncated to 400 words.
From these we derive temporal NLI features: running mean of contradiction, first-order difference between adjacent sentences, windowed maximum over 10 tokens, and an entailment drop score.

\paragraph{LM features (6 dimensions).}
We use TinyLlama-1.1B~\cite{zhang2024tinyllama} as an external observer.
For each token we extract log-probability (summed over constituent BPE subwords), next-token entropy, mean and maximum subword rank, a binary fallback indicator for vocabulary mismatches, and an interaction feature $\log P(w_t) \times \mathbf{1}[\text{matched}]$.
Missing values due to tokenizer mismatch are replaced with training-set medians to prevent test-set leakage.

Each signal family captures a different aspect of hallucination: text features detect divergence from the source; NLI features detect semantic contradiction; LM features detect statistical anomalies.
The temporal enrichment (running means, deltas, windowed extremes) makes sequential dynamics explicit in the feature representation itself, before the sequence model sees them.

\subsection{Sequence Labeling Models}
\label{sec:models}

Table~\ref{tab:architectures} summarizes the eight architectures we compare, spanning a spectrum from no temporal modeling to structured prediction.

\begin{table}[t]
\centering
\caption{Sequence labeling architectures. All models use $h{=}64$. CRF variants use forward-backward marginals for ranking (Section~\ref{sec:crf_results}).}
\label{tab:architectures}
\small
\begin{tabular}{@{}llr@{}}
\toprule
\textbf{Model} & \textbf{Temporal scope} & \textbf{Params} \\
\midrule
LogReg & None (per-token) & 34 \\
MLP & None (per-token) & 112K \\
1D-CNN & Local (kernel $\leq 7$) & 61K \\
BiGRU & Full sequence & 121K \\
BiLSTM & Full sequence & 121K \\
Transformer & Full sequence (attention) & 106K \\
BiGRU-CRF & Full + transitions & 121K \\
BiLSTM-CRF & Full + transitions & 121K \\
\bottomrule
\end{tabular}
\end{table}

The MLP baseline ($33 \to 384 \to 256 \to 1$, 112K parameters) matches BiGRU's parameter count to test whether sequential models win through capacity or cross-token modeling~\cite{kossen2024seps}.
BiGRU and BiLSTM are two-layer bidirectional networks with $h{=}64$ and dropout 0.1; the 1D-CNN uses dilated convolutions with kernels 3, 5, and 7.
CRF variants augment the base model with a Conditional Random Field output layer~\cite{lafferty2001crf} that models pairwise label transitions.

\subsection{Training}
\label{sec:training}

Algorithm~\ref{alg:training} summarizes the training procedure.
All neural models use AdamW~\cite{loshchilov2019adamw} with learning rate $10^{-3}$ and weight decay $10^{-4}$, following the hyperparameter recommendations for BiLSTM sequence labeling from \citet{reimers2017optimal}.
We train for up to 15 epochs with early stopping on validation F1 (patience 5) and gradient clipping at norm 1.0.
Sequences longer than 512 tokens are truncated (covering 97\% of RAGTruth examples); batch size is 32.
CRF models replace $\mathcal{L}$ with negative conditional log-likelihood.
A hyperparameter sweep over hidden dimensions $h \in \{32, 64, 128, 256\}$ and layers $\in \{1, 2, 3\}$ confirms that performance plateaus at $h{=}64$ and degrades for $h \geq 128$ due to the low input dimensionality (33 features); larger models overfit without improving AUC (Appendix~\ref{app:sweep}).
All experiments run on 2 RTX 4090 GPUs; total compute is under 8 GPU-hours.
Code is implemented in PyTorch 2.1.
Code and trained models will be made publicly available at \url{https://github.com/YehudaItkin/temporal-hallucination-detection} upon publication.

\begin{algorithm}[t]
\DontPrintSemicolon
\KwIn{Training set $\mathcal{D} = \{(\mathbf{X}^{(i)}, \mathbf{y}^{(i)})\}_{i=1}^{N}$, validation set $\mathcal{V}$}
\KwOut{Trained model $f_{\theta^*}$}
Compute class weight $\alpha \gets n_{\text{neg}} / n_{\text{pos}}$\;
Initialize $f_\theta$, optimizer $\gets$ AdamW($\theta$, lr=$10^{-3}$, wd=$10^{-4}$)\;
$\text{best\_F1} \gets 0$; \quad $\text{wait} \gets 0$\;
\For{epoch $= 1, \dots, 15$}{
  \For{batch $(\mathbf{X}, \mathbf{y}) \in \mathcal{D}$}{
    $\hat{\mathbf{y}} \gets f_\theta(\mathbf{X})$\;
    $\mathcal{L} \gets -\frac{1}{T}\sum_t [\alpha \, y_t \log \hat{y}_t + (1{-}y_t)\log(1{-}\hat{y}_t)]$ \tcp*{Eq.~\ref{eq:loss}}
    Backpropagate $\mathcal{L}$; clip $\|\nabla\| \leq 1.0$; step optimizer\;
  }
  $\text{F1}_\text{val} \gets \text{evaluate}(f_\theta, \mathcal{V})$\;
  \eIf{$\text{F1}_\text{val} > \text{best\_F1}$}{
    $\text{best\_F1} \gets \text{F1}_\text{val}$; \quad $\theta^* \gets \theta$; \quad $\text{wait} \gets 0$\;
  }{
    $\text{wait} \gets \text{wait} + 1$; \quad \lIf{wait $\geq 5$}{\textbf{break}}
  }
}
\Return $f_{\theta^*}$\;
\caption{Training with early stopping.}
\label{alg:training}
\end{algorithm}

\section{Experimental Setup}
\label{sec:experiments}

We design experiments to answer the three research questions stated in Section~\ref{sec:introduction}.
Table~\ref{tab:experiment_design} maps each research question to the experiments that address it.

\begin{table}[t]
\centering
\small
\caption{Research questions and corresponding experiments.}
\label{tab:experiment_design}
\begin{tabular}{@{}ll@{}}
\toprule
\textbf{RQ} & \textbf{Experiments} \\
\midrule
RQ1: Temporal & Main ablation (Section~\ref{sec:main_ablation}), \\
dependencies & decomposition (\ref{sec:decomposition_results}), \\
 & directional ablation (\ref{sec:directional_results}) \\
\addlinespace
RQ2: Multi-signal & Signal ablation (Section~\ref{sec:main_ablation}), \\
fusion & pointwise baselines (\ref{sec:pointwise_results}), \\
 & Lookback comparison (\ref{sec:lookback_results}) \\
\addlinespace
RQ3: General- & Cross-model (\ref{sec:cross_model_results}), \\
ization & cross-dataset (\ref{sec:cross_dataset_results}) \\
\bottomrule
\end{tabular}
\end{table}

\subsection{Datasets}
\label{sec:datasets}

\paragraph{RAGTruth}~\cite{niu2024ragtruth} is our primary benchmark.
It contains 15,090 training and 2,700 test examples spanning three task types: question answering (QA), summarization, and data-to-text generation.
Outputs are produced by six LLMs: GPT-4, GPT-3.5-turbo, LLaMA-2-Chat (7B, 13B, 70B)~\cite{touvron2023llama2}, and Mistral-7B-Instruct~\cite{jiang2023mistral}.
Span-level annotations distinguish evident conflict (contradicts the source) from baseless information (unsupported).
We convert span annotations to word-level binary labels: a word is hallucinated if more than 50\% of its character span overlaps with an annotated span.
The resulting dataset has a 5.6\% token-level hallucination rate.

\paragraph{PsiloQA}~\cite{psiloqa2025} is the transfer target.
PsiloQA is a multilingual span-level hallucination benchmark; we use its English QA subset, comprising 5,000 training, 1,098 test, and 890 validation examples spanning 11 source models, with a token-level hallucination prevalence of 53\%.
The ten-fold difference in positive rate versus RAGTruth creates a challenging domain shift.

\subsection{Experimental Protocol}
\label{sec:protocol}

We split RAGTruth's training set 85\%/15\% into train and validation subsets, stratified by the binary hallucination flag.
All random operations use a fixed seed (seed 42 for single-seed experiments; seeds 0--9 for multi-seed).
Feature extraction is deterministic (no learned parameters), so only the classifier varies across seeds.

\paragraph{Signal ablation.}
Four feature configurations (text only, text + NLI, text + LM, and all signals) are crossed with eight model architectures, yielding 32 conditions.

\paragraph{Cross-model transfer.}
Leave-one-out over the six source LLMs: train on five, test on the held-out model.

\paragraph{Cross-dataset transfer.}
Bidirectional zero-shot transfer between RAGTruth and PsiloQA, with no fine-tuning on the target domain.

\paragraph{Lookback Lens comparison.}
To contextualize our black-box approach, we compare against Lookback Lens~\cite{chuang2024lookback} on the open-source LLM subset of RAGTruth (10,060 train / 1,800 test).
Since the original method requires the generator's attention maps, we use proxy encoders: TinyLlama-1.1B (128-dim) and Qwen3-14B (160-dim).

\paragraph{Statistical testing.}
All primary comparisons use paired Wilcoxon signed-rank tests across 10 seeds.
With $n = 10$, the minimum achievable two-sided $p$-value is 0.002.
We do not apply multiple testing correction because our primary claim involves a single pre-specified comparison (BiGRU vs.\ LogReg); remaining analyses are exploratory.

\section{Results}
\label{sec:results}

\subsection{Main Ablation (RQ1, RQ2)}
\label{sec:main_ablation}

Table~\ref{tab:ablation} presents the core ablation across architectures and signal configurations.

\begin{table}[t]
\centering
\caption{Token-level AUC and F1 on RAGTruth test set. CRF models use softmax scoring; see Table~\ref{tab:crf} for forward-backward results.}
\label{tab:ablation}
\small
\begin{tabular}{lcccc}
\toprule
\multirow{2}{*}{\textbf{Model}} & \multicolumn{2}{c}{\textbf{Text Only}} & \multicolumn{2}{c}{\textbf{All Signals}} \\
\cmidrule(lr){2-3} \cmidrule(lr){4-5}
 & AUC & F1 & AUC & F1 \\
\midrule
LogReg         & 0.702 & 0.133 & 0.730 & 0.141 \\
1D-CNN         & 0.788 & 0.208 & 0.807 & 0.216 \\
BiLSTM         & 0.826 & 0.227 & 0.843 & \textbf{0.247} \\
\textbf{BiGRU} & \textbf{0.832} & 0.232 & \textbf{0.845} & 0.242 \\
Transformer    & 0.802 & 0.233 & 0.804 & 0.229 \\
\midrule
BiGRU-CRF      & 0.806 & 0.196 & 0.630 & 0.245 \\
BiLSTM-CRF     & 0.603 & 0.188 & 0.781 & 0.164 \\
Trans-CRF      & 0.668 & 0.235 & 0.691 & 0.227 \\
\bottomrule
\end{tabular}
\end{table}

Addressing RQ1, temporal models consistently outperform independent baselines.
BiGRU achieves 0.845 AUC with all signals (single seed); multi-seed analysis confirms $0.840 \pm 0.007$ (Table~\ref{tab:robustness}), exceeding LogReg ($0.730 \pm 0.001$) by 11.0 points ($p = 0.002$).
The gap is consistent across every signal configuration: even with text features alone, BiGRU exceeds LogReg by over 10 points.

Addressing RQ2, combining all three signal types consistently improves over any single source.
For BiGRU, adding NLI and LM features yields +1.3 points (0.832 to 0.845).
The effect is larger for weaker models: LogReg gains +2.8 points, 1D-CNN gains +1.9 points.

CRF-augmented models show degraded AUC under softmax scoring.
This is a scoring artifact, not a model deficiency, as we show in Section~\ref{sec:crf_results}.

\subsection{Robustness Across Seeds (RQ1)}
\label{sec:robustness_results}

\begin{table}[t]
\centering
\caption{Mean $\pm$ std across 10 seeds (all signals). MLP is a capacity-matched token-independent baseline (112K parameters $\approx$ BiGRU's 121K).}
\label{tab:robustness}
\small
\setlength{\tabcolsep}{4pt}
\begin{tabular}{@{}lccc@{}}
\toprule
\textbf{Model} & \textbf{AUC} & \textbf{F1} & \textbf{AP} \\
\midrule
LogReg      & $.730{\pm}.001$ & $.141{\pm}.000$ & $.093{\pm}.001$ \\
MLP         & $.765{\pm}.002$ & $.176{\pm}.003$ & $.127{\pm}.002$ \\
1D-CNN      & $.813{\pm}.002$ & $.218{\pm}.004$ & $.197{\pm}.005$ \\
BiLSTM      & $.835{\pm}.008$ & $.242{\pm}.008$ & $.270{\pm}.017$ \\
\textbf{BiGRU} & $\mathbf{.840{\pm}.007}$ & $\mathbf{.257{\pm}.010}$ & $\mathbf{.282{\pm}.018}$ \\
Transformer & $.819{\pm}.008$ & $.246{\pm}.009$ & $.234{\pm}.012$ \\
\bottomrule
\end{tabular}
\end{table}

All results are stable across seeds (standard deviations below 0.01 for five of six architectures; see Appendix~\ref{app:robustness_fig} for the full distribution).
BiGRU significantly outperforms every other architecture (all pairwise Wilcoxon $p < 0.05$).
The MLP baseline ($0.765 \pm 0.002$) closes only 32\% of the LogReg-to-BiGRU gap despite matching BiGRU's parameter count. The advantage comes from cross-token modeling, not nonlinear capacity.

The architecture itself is not the bottleneck.
The lead above reflects other models being undertrained under a fixed recipe rather than a genuine architectural gap: with plateau-based scheduling and early stopping, seven sequence models, including a Mamba state-space model and a bidirectional xLSTM, converge to an indistinguishable $0.843$--$0.845$ AUC, the largest gains going to the slowest-converging architectures (Appendix~\ref{app:arch_comparison}).
A ceiling shared across recurrent, state-space, and attention-augmented models indicates that the 33-dimensional feature set, not the architecture, limits performance.
We report BiGRU as the simplest model at this ceiling and the fastest to converge (epoch 8 versus 17--46 for the others), making it the practical default.

\subsection{Decomposition of the Temporal Advantage (RQ1)}
\label{sec:decomposition_results}

Four models form a controlled decomposition of BiGRU's 11-point advantage over LogReg:

\begin{table}[h]
\centering
\caption{Controlled decomposition of BiGRU's advantage over LogReg (10 seeds, all signals).}
\label{tab:decomposition}
\small
\begin{tabular}{@{}llcc@{}}
\toprule
\textbf{Model} & \textbf{Adds} & \textbf{AUC} & \textbf{$\Delta$} \\
\midrule
LogReg          & Per-token, linear    & .730 & -- \\
MLP (112K)      & Nonlinear capacity   & .765 & +3.5 (32\%) \\
Shuffled BiGRU  & Seq.\ aggregation    & .791 & +2.6 (24\%) \\
BiGRU           & Temporal order       & .840 & +4.9 (44\%) \\
\bottomrule
\end{tabular}
\end{table}

\begin{figure}[t]
\centering
\includegraphics[width=0.85\linewidth]{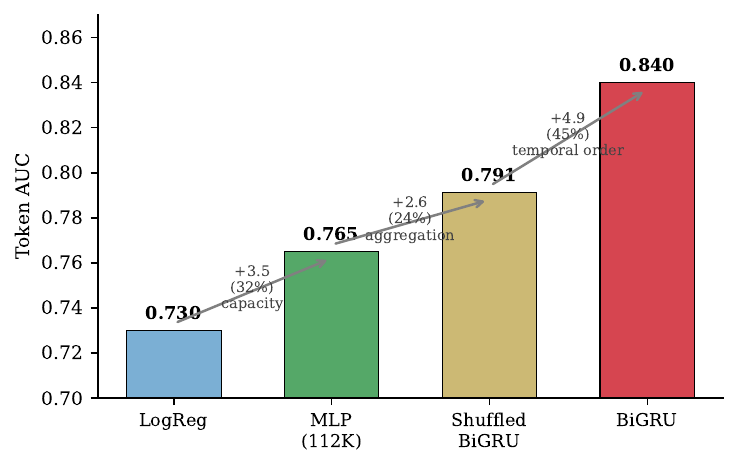}
\caption{Decomposition of BiGRU's 11-point advantage over LogReg. Temporal order is the largest contributor (44\%).}
\label{fig:decomposition}
\end{figure}

The shuffled BiGRU is trained and evaluated on randomly permuted token sequences (features and labels in lockstep), measuring what BiGRU can do as a nonlinear set aggregator without temporal structure.
The 4.9-point gain from temporal order ($p = 0.001$, 10 seeds) is the largest single component (Figure~\ref{fig:decomposition}).

\subsection{Directional Ablation (RQ1)}
\label{sec:directional_results}

\begin{table}[h]
\centering
\caption{Directional ablation (10 seeds, all signals).}
\label{tab:directional}
\small
\begin{tabular}{lcc}
\toprule
\textbf{Model} & \textbf{AUC} & \textbf{Avg Precision} \\
\midrule
ForwardGRU   & $.802 \pm .008$ & $.217 \pm .014$ \\
BackwardGRU  & $.817 \pm .005$ & $.223 \pm .015$ \\
BiGRU        & $\mathbf{.840 \pm .007}$ & $\mathbf{.282 \pm .018}$ \\
\bottomrule
\end{tabular}
\end{table}

Both unidirectional models outperform the shuffled BiGRU (0.791). Token order carries information in both directions.
BackwardGRU outperforms ForwardGRU by 1.5 points ($p < 0.001$) because our features already encode forward-looking statistics (running means, deltas), making the forward recurrence partially redundant.
The backward direction provides complementary information about the context following a hallucinated region.
BiGRU captures both and outperforms each unidirectional model ($p < 0.001$).

\subsection{Pointwise Baselines (RQ2)}
\label{sec:pointwise_results}

\begin{table}[t]
\centering
\caption{Individual features as standalone detectors.}
\label{tab:pointwise}
\small
\begin{tabular}{llcc}
\toprule
\textbf{Signal} & \textbf{Method class} & \textbf{AUC} & \textbf{AP} \\
\midrule
LM entropy          & Token entropy     & .551 & .047 \\
LM log-prob         & Perplexity        & .549 & .046 \\
NLI contradiction   & NLI-based         & .639 & .087 \\
NLI entailment drop & Entailment check  & .641 & .066 \\
Context overlap     & Lexical overlap   & .597 & .054 \\
Running novelty     & Novelty tracking  & .616 & .055 \\
\midrule
\multicolumn{2}{l}{LogReg, all 33 features} & .730 & .103 \\
\multicolumn{2}{l}{\textbf{BiGRU, all 33 features}} & $\mathbf{.840}$ & $\mathbf{.282}$ \\
\bottomrule
\end{tabular}
\end{table}

The best single feature (NLI entailment drop, 0.641 AUC) falls 20 points below BiGRU.
LM entropy reaches only 0.551 (barely above chance) because we use a proxy model (TinyLlama), not the generator itself.
The 20-point gap decomposes into multi-signal fusion (+8.9 points, individual to LogReg) and temporal modeling (+11.0 points, LogReg to BiGRU).

\subsection{Information-Theoretic Analysis (RQ1)}
\label{sec:analysis}

The results above establish that temporal modeling helps; we now ask \emph{why}.
We model the hallucination label sequence as a Markov chain with transition matrix
\begin{equation}
\mathbf{T} = \begin{pmatrix} 0.994 & 0.006 \\ 0.098 & 0.902 \end{pmatrix}
\end{equation}
and stationary distribution $\boldsymbol{\pi} = (0.942, 0.058)$, close to the observed 5.6\% rate.
The entropy rate $h = 0.077$ bits/token measures uncertainty \emph{after} observing the previous label; the marginal entropy $H(y_t) = 0.318$ bits measures uncertainty without context.
Their ratio gives the temporal redundancy: $1 - h / H(y_t) = 76\%$ --- three-quarters of the uncertainty about whether a token is hallucinated can be resolved by knowing the previous token's label.
An independent classifier discards this information entirely.

Conditioning on both neighbors reduces uncertainty further: $H(y_t \mid y_{t-1}, y_{t+1}) = 0.029$ bits, meaning bidirectional context captures 91\% of label entropy.
This is consistent with our directional ablation: BiGRU gains +3.8 AUC over ForwardGRU and +2.3 over BackwardGRU.
The mutual information $I(y_t; y_{t-k})$ decays geometrically with lag $k$ at rate $\lambda_2 = 0.896$, giving a mixing time of $\sim$10 tokens.
This predicts that local models (1D-CNN, kernel $\leq 7$) should capture most temporal signal: they account for 74\% of theoretical MI and recover 75\% of the BiGRU--LogReg gap empirically (Appendix~\ref{app:span_evolution}).
A counterfactual analysis confirms that NLI features discriminate genuine hallucination while LM features partly detect generator style (Appendix~\ref{app:counterfactual}).

\subsection{Cross-Model Generalization (RQ3)}
\label{sec:cross_model_results}

Leave-one-out transfer across six source LLMs yields a mean AUC of 0.808, a 3.8\% relative degradation from in-distribution (0.840).
Performance varies by target: LLaMA-2-70B is easiest (0.829), GPT-4 is hardest (0.781), likely due to its lower hallucination rate and qualitatively different error patterns.
The temporal advantage persists: BiGRU consistently outperforms LogReg and Transformer across all six held-out models.

\subsection{Cross-Dataset Transfer (RQ3)}
\label{sec:cross_dataset_results}

\begin{table}[t]
\centering
\caption{Cross-dataset transfer (mean $\pm$ std, 10 seeds, all signals).}
\label{tab:transfer}
\small
\begin{tabular}{lcccc}
\toprule
\multirow{2}{*}{\textbf{Model}} & \multicolumn{2}{c}{\textbf{R $\to$ P}} & \multicolumn{2}{c}{\textbf{P $\to$ R}} \\
\cmidrule(lr){2-3} \cmidrule(lr){4-5}
 & AUC & F1 & AUC & F1 \\
\midrule
LogReg      & $.618$ & $.712$ & $.692$ & $.161$ \\
BiGRU       & $.634$ & $.709$ & $\mathbf{.744}$ & $.185$ \\
Transformer & $.624$ & $.719$ & $.703$ & $.144$ \\
\bottomrule
\end{tabular}
\end{table}

Transfer is asymmetric.
Training on PsiloQA (4,250 training examples after our 85/15 split, 53\% positive) and testing on RAGTruth achieves 0.744 AUC, outperforming the reverse direction (0.634) despite having three times fewer training examples.
We attribute this to annotation density: PsiloQA's 4,250 examples contain roughly three times more hallucinated tokens than RAGTruth's 12,826 examples (53\% vs.\ 5.6\% positive rate).
F1 values across the two directions are not directly comparable because they reflect different label distributions in the respective test sets.

\subsection{Lookback Lens Comparison}
\label{sec:lookback_results}

\begin{table}[t]
\centering
\caption{Fair comparison on the open-LLM subset of RAGTruth (5 seeds). Combined = our 33-dim + Lookback features.}
\label{tab:lookback}
\small
\begin{tabular}{llcc}
\toprule
\textbf{Features} & \textbf{Model} & \textbf{AUC} & \textbf{AP} \\
\midrule
\multicolumn{4}{l}{\emph{Proxy: Qwen3-14B}} \\
Lookback only       & BiGRU  & $.838 \pm .015$ & $.368$ \\
Combined (193-dim)  & BiGRU  & $\mathbf{.866 \pm .007}$ & $\mathbf{.425}$ \\
\midrule
\multicolumn{4}{l}{\emph{No attention access}} \\
Ours only (33-dim)  & BiGRU  & $.819 \pm .013$ & $.298$ \\
\bottomrule
\end{tabular}
\end{table}

On the open-source subset, Lookback Lens with Qwen3-14B proxy achieves 0.838 AUC, slightly above our features alone (0.819).
Combining both feature sets yields 0.866 AUC (+2.8 over Lookback alone). The two signal types capture orthogonal aspects of hallucination.
Our approach remains the only one applicable to closed-source LLMs, where it achieves 0.840 AUC on the full benchmark.

\subsection{CRF Calibration}
\label{sec:crf_results}

\begin{table}[t]
\centering
\caption{CRF scoring: softmax vs.\ forward-backward (FB) marginals. Softmax values differ from Table~\ref{tab:ablation} because CRF softmax scores are unstable across seeds (see text).}
\label{tab:crf}
\small
\begin{tabular}{lccc}
\toprule
\textbf{Model} & \textbf{Softmax} & \textbf{FB} & \textbf{$\Delta$} \\
\midrule
BiGRU-CRF  & 0.666 & \textbf{0.845} & +0.179 \\
BiLSTM-CRF & 0.785 & 0.807 & +0.022 \\
Trans-CRF  & 0.741 & 0.818 & +0.077 \\
\bottomrule
\end{tabular}
\end{table}

With forward-backward marginals, BiGRU-CRF recovers from 0.666 to 0.845 AUC, matching the non-CRF BiGRU.
The issue is that CRF emission logits are not calibrated probabilities; they are potentials whose interpretation depends on the transition matrix.
Softmax ignores this structure and produces scores that vary by up to $\pm 0.05$ across training runs.
Forward-backward marginals account for the full transition structure and produce stable rankings.
This finding extends to any setting where CRF models are evaluated with ranking metrics.

\section{Discussion}
\label{sec:discussion}

\subsection{Interpreting the Temporal Advantage}
\label{sec:interpreting}

Existing detectors classify each token independently, and our results quantify what that assumption costs: 11 AUC points relative to a sequence model trained on identical features, with 44\% of the gap attributable to token order alone (Section~\ref{sec:decomposition_results}).
The information-theoretic analysis of Section~\ref{sec:analysis} explains where this gain comes from.
With 76\% of label entropy resolved by one step of context and 91\% by both neighbors, an independent classifier discards most of what is knowable about a token's label before its features are even consulted; a bidirectional model recovers that information.

The same analysis bounds how far the temporal signal reaches: with a mixing time of roughly 10 tokens, the advantage is local rather than long-range, and the success of the local 1D-CNN (Section~\ref{sec:analysis}) bears this out.
The design implication is that when inference latency matters, a local convolutional model is a sound default; the recurrent model buys the last quarter of the gain.

The comparison also separates two senses of ``temporal.''
Our input features already include running means, deltas, and windowed extremes, so the BiGRU vs.\ LogReg gap measures whether modeling \emph{interactions between positions} adds value beyond per-token temporal summaries; the 11-point gap confirms it does.
The same observation explains the directional asymmetry of Section~\ref{sec:directional_results}: the features encode forward-looking statistics, which makes the forward recurrence partially redundant, while the backward direction contributes complementary information about the context that \emph{follows} a hallucinated region.
Appendix~\ref{app:span_evolution} visualizes the mechanism: BiGRU maintains elevated hallucination probability throughout spans, while ForwardGRU rises most sharply at onset.

Finally, the fused signals are not interchangeable: the counterfactual analysis of Appendix~\ref{app:counterfactual} cautions against relying on surprisal alone, since LM features partly track generator style rather than hallucination itself, and explains why fusion, not any single family, drives the result.

\subsection{Practical Considerations}
\label{sec:practical}

Our BiGRU achieves 0.840 token-level AUC but 0.394 span-level F1, compared to LettuceDetect's 0.589 span F1.
This gap reflects different optimization targets and architectural scales, not a failure of temporal modeling.
LettuceDetect fine-tunes a 395M-parameter encoder for token classification, learning contextual representations that produce sharp span boundaries.
Our approach uses 121K parameters on pre-extracted features, optimized for ranking quality (AUC).
Token F1 further penalizes threshold miscalibration under 5.6\% positive rate.
Closing the span F1 gap would likely require replacing pre-extracted features with a learned encoder, at the cost of black-box applicability and a $3000\times$ increase in parameters.

Three additional findings have direct implications for deployment.
First, annotation density matters more than dataset size for training transferable detectors: a small, densely annotated dataset (PsiloQA, 53\% positive, 4,250 examples) transfers better than a large, sparse one (RAGTruth, 5.6\%, 12,826 examples).
This suggests that practitioners building custom detectors should prioritize label quality over volume.
Second, the black-box nature of our approach makes it the only method applicable to closed-source APIs, which dominate production usage.
Third, combining our features with attention-based signals when white-box access is available yields 0.866 AUC, suggesting a two-tier deployment strategy: black-box features as the default, augmented with attention features when the model is open-source.

\section{Conclusion}
\label{sec:conclusion}

We framed token-level hallucination detection as sequence labeling over a multi-signal feature stream and showed that temporal structure is the largest untapped signal in the task: a BiGRU exceeds independent classifiers by 11 AUC points on identical features, and a controlled decomposition attributes the largest share of that gain to token order itself.

Three lessons generalize beyond our setting.
Multi-signal fusion is necessary but not sufficient: without temporal modeling, fusion stalls at 0.730 AUC.
Annotation density matters more than dataset size for transfer, so practitioners should prioritize label quality over volume.
And boundary precision trades off against accessibility: closing the span-F1 gap to fine-tuned encoders would require white-box access and three orders of magnitude more parameters.

Because the detector needs no access to the generating model, degrades by under 4\% on unseen source models, and combines additively with attention-based signals when those are available, it is directly deployable against closed-source APIs.
Future work should explore streaming detection during generation, where a unidirectional variant could trigger intervention before a hallucinated span completes.

\section*{Limitations}
Our approach has several limitations.
Feature extraction requires inference passes through DeBERTa (350M parameters) and TinyLlama (1.1B parameters) at test time; in latency-sensitive applications, this overhead may be a limiting factor.
All experiments use English-language datasets; hallucination patterns may differ in morphologically rich languages.
The method is black-box by design, which is a practical necessity for closed-source LLMs but means we do not exploit internal signals that could improve detection when available.
Our binary formulation does not distinguish hallucination severity or type.
Token-level AUC treats tokens within the same example as independent observations, which inflates effective sample size; example-level AUC (0.795 for BiGRU) is consistently lower but preserves model rankings.
RAGTruth's 5.6\% positive rate limits token-level F1 despite strong AUC.
NLI features truncate source contexts to 400 words, potentially missing contradictions with later content.
Our span-level F1 (0.394) is substantially below LettuceDetect (0.589). This gap reflects optimization for ranking quality rather than boundary precision, at 3000$\times$ fewer parameters.

\section*{Ethics Statement}
This work develops tools for detecting hallucinated content in LLM outputs.
All datasets used (RAGTruth, PsiloQA) are publicly available benchmarks containing no personally identifiable information.
We acknowledge that detection signals could theoretically help adversaries craft harder-to-detect hallucinations, but we believe the benefit of improved detection outweighs this risk.

\section*{Acknowledgements}
We thank Ilya Makarov for detailed feedback on the manuscript and Professor Ivan Oseledets for his valuable comments.

\bibliography{references}

\begin{thebibliography}{40}
\providecommand{\natexlab}[1]{#1}

\bibitem[{Akarlar(2026)}]{akarlar2026trajectory}
G.~Aytug Akarlar. 2026.
\newblock Hallucination as trajectory commitment: Causal evidence for
  asymmetric attractor dynamics in transformer generation.
\newblock \emph{arXiv preprint arXiv:2604.15400}.

\bibitem[{Chatterjee et~al.(2025)Chatterjee, Goel, and Chakraborty}]{hide2025}
Anwoy Chatterjee, Yash Goel, and Tanmoy Chakraborty. 2025.
\newblock {HIDE} and seek: Detecting hallucinations in language models via
  decoupled representations.
\newblock \emph{arXiv preprint arXiv:2506.17748}.

\bibitem[{Cheang et~al.(2025)Cheang, Chan, Zhang, and
  Deng}]{cheng2025llmsdonotknow}
Chi~Seng Cheang, Hou~Pong Chan, Wenxuan Zhang, and Yang Deng. 2025.
\newblock Do {LLMs} really know what they don't know? internal states mainly
  reflect knowledge recall rather than truthfulness.
\newblock \emph{arXiv preprint arXiv:2510.09033}.

\bibitem[{Chen et~al.(2026)Chen, Fan, Wang, Leng, Wu, Zheng, Sun, and
  Wu}]{hallusae2026}
Boshui Chen, Zhaoxin Fan, Ke~Wang, Zhiying Leng, Faguo Wu, Hongwei Zheng, Yifan
  Sun, and Wenjun Wu. 2026.
\newblock {HalluSAE}: Detecting hallucinations in large language models via
  sparse auto-encoders.
\newblock \emph{arXiv preprint arXiv:2604.16430}.

\bibitem[{Chen et~al.(2024)Chen, Liu, Chen, Gu, Wu, Tao, Fu, and
  Ye}]{chen2024inside}
Chao Chen, Kai Liu, Ze~Chen, Yi~Gu, Yue Wu, Mingyuan Tao, Zhihang Fu, and
  Jieping Ye. 2024.
\newblock {INSIDE}: {LLM}'s internal states retain the power of hallucination
  detection.
\newblock In \emph{International Conference on Learning Representations}.

\bibitem[{Chuang et~al.(2024)Chuang, Qiu, Hsieh, Krishna, Kim, and
  Glass}]{chuang2024lookback}
Yung-Sung Chuang, Linlu Qiu, Cheng-Yu Hsieh, Ranjay Krishna, Yoon Kim, and
  James Glass. 2024.
\newblock Lookback lens: Detecting and mitigating contextual hallucinations in
  large language models using only attention maps.
\newblock In \emph{Proceedings of the 2024 Conference on Empirical Methods in
  Natural Language Processing}, pages 1438--1451.

\bibitem[{Dubanowska et~al.(2025)Dubanowska, {\.Z}elaszczyk, Brzozowski,
  Mandica, and Karpowicz}]{dubanowska2025oodgeneralization}
Zuzanna Dubanowska, Maciej {\.Z}elaszczyk, Micha{\l} Brzozowski, Paolo Mandica,
  and Micha{\l} Karpowicz. 2025.
\newblock \href {https://arxiv.org/abs/2509.19372} {Representation-based broad
  hallucination detectors fail to generalize out of distribution}.
\newblock In \emph{Findings of EMNLP}.

\bibitem[{Fadeeva et~al.(2023)Fadeeva, Vashurin, Tsvigun, Vazhentsev, Petrakov,
  Fedyanin, Vasilev, Goncharova, Panchenko, Panov, Baldwin, and
  Shelmanov}]{fadeeva2024lmpolygraph}
Ekaterina Fadeeva, Roman Vashurin, Akim Tsvigun, Artem Vazhentsev, Sergey
  Petrakov, Kirill Fedyanin, Daniil Vasilev, Elizaveta Goncharova, Alexander
  Panchenko, Maxim Panov, Timothy Baldwin, and Artem Shelmanov. 2023.
\newblock \href {https://arxiv.org/abs/2311.07383} {{LM-Polygraph}: Uncertainty
  estimation for language models}.
\newblock In \emph{Proceedings of EMNLP: System Demonstrations}.

\bibitem[{Farquhar et~al.(2024)Farquhar, Kossen, Kuhn, and
  Gal}]{farquhar2024semanticentropy}
Sebastian Farquhar, Jannik Kossen, Lorenz Kuhn, and Yarin Gal. 2024.
\newblock Detecting hallucinations in large language models using semantic
  entropy.
\newblock \emph{Nature}, 630:625--630.

\bibitem[{Ferrando et~al.(2024)Ferrando, Obeso, Rajamanoharan, and
  Nanda}]{ferrando2024entityknowledge}
Javier Ferrando, Oscar Obeso, Senthooran Rajamanoharan, and Neel Nanda. 2024.
\newblock Do {I} know this entity? knowledge awareness and hallucinations in
  language models.
\newblock \emph{arXiv preprint arXiv:2411.14257}.

\bibitem[{He et~al.(2023)He, Gao, and Chen}]{he2023debertav3}
Pengcheng He, Jianfeng Gao, and Weizhu Chen. 2023.
\newblock {DeBERTaV3}: Improving {DeBERTa} using {ELECTRA}-style pre-training
  with gradient-disentangled embedding sharing.
\newblock In \emph{International Conference on Learning Representations}.

\bibitem[{He et~al.(2021)He, Liu, Gao, and Chen}]{he2021deberta}
Pengcheng He, Xiaodong Liu, Jianfeng Gao, and Weizhu Chen. 2021.
\newblock {DeBERTa}: Decoding-enhanced {BERT} with disentangled attention.
\newblock In \emph{International Conference on Learning Representations}.

\bibitem[{Huang et~al.(2023)Huang, Yu, Ma, Zhong, Feng, Wang, Chen, Peng, Feng,
  Qin, and Liu}]{huang2023survey}
Lei Huang, Weijiang Yu, Weitao Ma, Weihong Zhong, Zhangyin Feng, Haotian Wang,
  Qianglong Chen, Weihua Peng, Xiaocheng Feng, Bing Qin, and Ting Liu. 2023.
\newblock A survey on hallucination in large language models: Principles,
  taxonomy, challenges, and open questions.
\newblock \emph{arXiv preprint arXiv:2311.05232}.

\bibitem[{Ji et~al.(2023)Ji, Lee, Frieske, Yu, Su, Xu, Ishii, Bang, Madotto,
  and Fung}]{ji2023survey}
Ziwei Ji, Nayeon Lee, Rita Frieske, Tiezheng Yu, Dan Su, Yan Xu, Etsuko Ishii,
  Ye~Jin Bang, Andrea Madotto, and Pascale Fung. 2023.
\newblock Survey of hallucination in natural language generation.
\newblock \emph{ACM Computing Surveys}, 55(12):1--38.

\bibitem[{Jiang et~al.(2023)Jiang, Sablayrolles, Mensch
  et~al.}]{jiang2023mistral}
Albert~Q Jiang, Alexandre Sablayrolles, Arthur Mensch, and 1 others. 2023.
\newblock {Mistral} 7{B}.
\newblock \emph{arXiv preprint arXiv:2310.06825}.

\bibitem[{Kossen et~al.(2024)Kossen, Han, Razzak, Schut, Malik, and
  Gal}]{kossen2024seps}
Jannik Kossen, Jiatong Han, Muhammed Razzak, Lisa Schut, Shreshth Malik, and
  Yarin Gal. 2024.
\newblock Semantic entropy probes: Robust and cheap hallucination detection in
  {LLMs}.
\newblock \emph{arXiv preprint arXiv:2406.15927}.

\bibitem[{Kov{\'a}cs and Recski(2025)}]{lettucedetect2025}
{\'A}d{\'a}m Kov{\'a}cs and G{\'a}bor Recski. 2025.
\newblock Lettucedetect: A hallucination detection framework for {RAG}
  applications.
\newblock \emph{arXiv preprint arXiv:2502.17125}.

\bibitem[{Lafferty et~al.(2001)Lafferty, McCallum, and
  Pereira}]{lafferty2001crf}
John Lafferty, Andrew McCallum, and Fernando Pereira. 2001.
\newblock Conditional random fields: Probabilistic models for segmenting and
  labeling sequence data.
\newblock In \emph{International Conference on Machine Learning}.

\bibitem[{Loshchilov and Hutter(2019)}]{loshchilov2019adamw}
Ilya Loshchilov and Frank Hutter. 2019.
\newblock Decoupled weight decay regularization.
\newblock In \emph{International Conference on Learning Representations}.

\bibitem[{Luan et~al.(2026)Luan, Li, Qin, Guo, Zhou, Wu, Zheng, Wu, and
  Fan}]{luan2026lyapunov}
Bozhi Luan, Gen Li, Yalan Qin, Jifeng Guo, Yun Zhou, Faguo Wu, Hongwei Zheng,
  Wenjun Wu, and Zhaoxin Fan. 2026.
\newblock Lyapunov probes for hallucination detection in large foundation
  models.
\newblock \emph{arXiv preprint arXiv:2603.06081}.

\bibitem[{Min et~al.(2026)Min, Pang, Du, Cheng, and Fung}]{min2026tokenhd}
Rui Min, Tianyu Pang, Chao Du, Minhao Cheng, and Yi~R. Fung. 2026.
\newblock Scalable token-level hallucination detection in large language
  models.
\newblock \emph{arXiv preprint arXiv:2605.12384}.

\bibitem[{Moslonka et~al.(2025)Moslonka, Randrianarivo, Garnier, and
  Malherbe}]{entropyproduction2025}
Charles Moslonka, Hicham Randrianarivo, Arthur Garnier, and Emmanuel Malherbe.
  2025.
\newblock Learned hallucination detection in black-box {LLMs} using token-level
  entropy production rate.
\newblock \emph{arXiv preprint arXiv:2509.04492}.

\bibitem[{Niu et~al.(2024)Niu, Wu, Zhu, Xu, Shum, Zhong, Song, and
  Zhang}]{niu2024ragtruth}
Cheng Niu, Yuanhao Wu, Juno Zhu, Siliang Xu, KaShun Shum, Randy Zhong, Juntong
  Song, and Tong Zhang. 2024.
\newblock \href {https://arxiv.org/abs/2401.00396} {{RAGTruth}: A hallucination
  corpus for developing trustworthy retrieval-augmented language models}.
\newblock In \emph{Proceedings of ACL}.

\bibitem[{Obeso et~al.(2025)Obeso, Arditi, Ferrando, Freeman, Holmes, and
  Nanda}]{obeso2025realtime}
Oscar Obeso, Andy Arditi, Javier Ferrando, Joshua Freeman, Cameron Holmes, and
  Neel Nanda. 2025.
\newblock Real-time detection of hallucinated entities in long-form generation.
\newblock \emph{arXiv preprint arXiv:2509.03531}.

\bibitem[{O'Neill et~al.(2025)O'Neill, Chalnev, Zhao, Kirkby, and
  Jayasekara}]{singledirection2025}
Charles O'Neill, Slava Chalnev, Chi~Chi Zhao, Max Kirkby, and Mudith
  Jayasekara. 2025.
\newblock A single direction of truth: An observer model's linear residual
  probe exposes and steers contextual hallucinations.
\newblock \emph{arXiv preprint arXiv:2507.23221}.

\bibitem[{Reimers and Gurevych(2017)}]{reimers2017optimal}
Nils Reimers and Iryna Gurevych. 2017.
\newblock \href {https://arxiv.org/abs/1707.06799} {Optimal hyperparameters for
  deep {LSTM}-networks for sequence labeling tasks}.
\newblock In \emph{Proceedings of the 2017 Conference on Empirical Methods in
  Natural Language Processing}.

\bibitem[{Roy et~al.(2026)Roy, Misra, Singh, and Roy}]{roy2026detection}
Dip Roy, Rajiv Misra, Sanjay~Kumar Singh, and Anisha Roy. 2026.
\newblock Detection without correction: A robust asymmetry in activation-based
  hallucination probing.
\newblock \emph{arXiv preprint arXiv:2604.13068}.

\bibitem[{Rykov et~al.(2025)Rykov, Petrushina, Savkin, Olisov, Vazhentsev,
  Titova et~al.}]{psiloqa2025}
Elisei Rykov, Kseniia Petrushina, Maksim Savkin, Valerii Olisov, Artem
  Vazhentsev, Kseniia Titova, and 1 others. 2025.
\newblock \href {https://arxiv.org/abs/2510.04849} {When models lie, we learn:
  Multilingual span-level hallucination detection with {PsiloQA}}.
\newblock In \emph{Findings of EMNLP}.

\bibitem[{Shapiro et~al.(2026)Shapiro, Taneja, and Goel}]{shapiro2026halt}
Ahmad Shapiro, Karan Taneja, and Ashok Goel. 2026.
\newblock {HALT}: Hallucination assessment via log-probs as time series.
\newblock \emph{arXiv preprint arXiv:2602.02888}.

\bibitem[{Simhi et~al.(2025)Simhi, Itzhak, Barez, Stanovsky, and
  Belinkov}]{zhang2025trustmewrong}
Adi Simhi, Itay Itzhak, Fazl Barez, Gabriel Stanovsky, and Yonatan Belinkov.
  2025.
\newblock Trust me, {I'm} wrong: {LLMs} hallucinate with certainty despite
  knowing the answer.
\newblock \emph{arXiv preprint arXiv:2502.12964}.

\bibitem[{Singh et~al.(2026)Singh, Paudel, and Roy}]{ragtstacking2025}
Akshita Singh, Prabesh Paudel, and Siddhartha Roy. 2026.
\newblock Hallucination detection via activations of open-weight proxy
  analyzers.
\newblock \emph{arXiv preprint arXiv:2605.07209}.

\bibitem[{Snel and Oh(2025)}]{snel2025firsttokens}
Jakob Snel and Seong~Joon Oh. 2025.
\newblock First hallucination tokens are different from conditional ones.
\newblock \emph{arXiv preprint arXiv:2507.20836}.

\bibitem[{Su et~al.(2025)Su, Hu, Koppula, Krishna, Pouransari, Hsieh
  et~al.}]{ogasa2025rl4hs}
Hsuan Su, Ting-Yao Hu, Hema~Swetha Koppula, Kundan Krishna, Hadi Pouransari,
  Cheng-Yu Hsieh, and 1 others. 2025.
\newblock Learning to reason for hallucination span detection.
\newblock \emph{arXiv preprint arXiv:2510.02173}.

\bibitem[{Sun et~al.(2024)Sun, Zang, Zheng, Song, Xu, Zhang, Yu, and
  Li}]{redeep2024}
Zhongxiang Sun, Xiaoxue Zang, Kai Zheng, Yang Song, Jun Xu, Xiao Zhang, Weijie
  Yu, and Han Li. 2024.
\newblock {ReDeEP}: Detecting hallucination in retrieval-augmented generation
  via mechanistic interpretability.
\newblock \emph{arXiv preprint arXiv:2410.11414}.

\bibitem[{Touvron et~al.(2023)Touvron, Martin, Stone
  et~al.}]{touvron2023llama2}
Hugo Touvron, Louis Martin, Kevin Stone, and 1 others. 2023.
\newblock {Llama} 2: Open foundation and fine-tuned chat models.
\newblock \emph{arXiv preprint arXiv:2307.09288}.

\bibitem[{{vLLM Team}(2025)}]{halugate2025}
{vLLM Team}. 2025.
\newblock \href {https://blog.vllm.ai/2025/12/14/halugate.html} {{HaluGate}:
  Token-level truth for production {LLMs}}.
\newblock vLLM Blog.

\bibitem[{Wang et~al.(2026)Wang, Cao, Wilson, and Zeng}]{wang2026fepoid}
Xinpeng Wang, William Cao, Andrew~Gordon Wilson, and Zhe Zeng. 2026.
\newblock Automatic layer selection for hallucination detection.
\newblock \emph{arXiv preprint arXiv:2605.26366}.

\bibitem[{Williams et~al.(2018)Williams, Nangia, and Bowman}]{williams2018mnli}
Adina Williams, Nikita Nangia, and Samuel Bowman. 2018.
\newblock A broad-coverage challenge corpus for sentence understanding through
  inference.
\newblock In \emph{Proceedings of NAACL-HLT}.

\bibitem[{Xiong et~al.(2025)Xiong, He, Liu, Sinha, and Zhang}]{raglens2025}
Guangzhi Xiong, Zhenghao He, Bohan Liu, Sanchit Sinha, and Aidong Zhang. 2025.
\newblock Toward faithful retrieval-augmented generation with sparse
  autoencoders.
\newblock \emph{arXiv preprint arXiv:2512.08892}.

\bibitem[{Zhang et~al.(2024)Zhang, Zeng, Wang, and Lu}]{zhang2024tinyllama}
Peiyuan Zhang, Guangtao Zeng, Tianduo Wang, and Wei Lu. 2024.
\newblock {TinyLlama}: An open-source small language model.
\newblock \emph{arXiv preprint arXiv:2401.02385}.

\end{thebibliography}

\clearpage
\appendix
\section{Complete Feature List}
\label{app:features}

Table~\ref{tab:full_features} lists all 33 per-token features grouped by signal family.
Of these, 12 features encode temporal context (cumulative statistics, sliding windows, or finite differences), while 21 are pointwise.
Even with temporal enrichment available to all models, BiGRU adds +11 AUC points over LogReg by modeling \emph{interactions between positions} that per-token statistics cannot capture.

\begin{table}[H]
\centering
\small
\caption{Complete list of 33 per-token features. The ``Context'' column indicates whether the feature encodes temporal information: \textbf{point} = depends only on the current token; \textbf{cumul} = cumulative statistic up to position $t$; \textbf{window} = sliding window average/max; \textbf{delta} = finite difference between adjacent positions; \textbf{sent} = sentence-level (projected to all tokens in the sentence).}
\label{tab:full_features}
\setlength{\tabcolsep}{3pt}
\begin{tabular}{@{}clp{3.8cm}l@{}}
\toprule
\textbf{\#} & \textbf{Group} & \textbf{Feature} & \textbf{Context} \\
\midrule
1 & Text & Word length (/20) & point \\
2 & Text & Is numeric & point \\
3 & Text & Is capitalized & point \\
4 & Text & Absolute position $t$ & point \\
5 & Text & Relative position $t/(T{-}1)$ & point \\
6 & Text & Unigram overlap $\mathbf{1}[w_t \in C]$ & point \\
7 & Text & Bigram overlap & point \\
8 & Text & Trigram overlap & point \\
9 & Text & Entity indicator & point \\
10 & Text & Cumulative overlap ratio & \textbf{cumul} \\
11 & Text & Running novelty rate & \textbf{cumul} \\
12 & Text & Consecutive novel count & \textbf{cumul} \\
13 & Text & Windowed novelty (5 tok) & \textbf{window} \\
14 & Text & Windowed novelty (10 tok) & \textbf{window} \\
15 & Text & Windowed novelty (20 tok) & \textbf{window} \\
16 & Text & Novelty velocity (1st diff) & \textbf{delta} \\
17 & Text & Novelty accel.\ (2nd diff) & \textbf{delta} \\
18 & Text & Sentence position index & point \\
19 & Text & Sentence-relative position & point \\
20 & Text & Running mean word length & \textbf{cumul} \\
\midrule
21 & NLI & $P(\text{contradiction} \mid C, s)$ & sent \\
22 & NLI & $P(\text{entailment} \mid C, s)$ & sent \\
23 & NLI & $P(\text{neutral} \mid C, s)$ & sent \\
24 & NLI & Running mean contradiction & \textbf{cumul} \\
25 & NLI & Contradiction delta & \textbf{delta} \\
26 & NLI & Windowed max contr.\ (10 tok) & \textbf{window} \\
27 & NLI & Entailment drop & sent \\
\midrule
28 & LM & Log-probability (sum subwords) & point \\
29 & LM & Next-token entropy & point \\
30 & LM & Mean subword rank (log) & point \\
31 & LM & Max subword rank (log) & point \\
32 & LM & Fallback indicator & point \\
33 & LM & Interaction: $\log P \times$ matched & point \\
\bottomrule
\end{tabular}
\end{table}

\section{Preprocessing Details}
\label{app:preprocessing}

We convert RAGTruth's span-level annotations to word-level binary labels.
A word is labeled hallucinated if more than 50\% of its character span overlaps with an annotated hallucination span.
NLI scores are computed at the sentence level using DeBERTa-v3-large~\cite{he2023debertav3} fine-tuned on Multi-Genre NLI~\cite{williams2018mnli}, with the source context truncated to 400 words to fit the 512-token encoder limit.
LM features use TinyLlama-1.1B~\cite{zhang2024tinyllama} as a proxy observer; tokens with no matching BPE subword receive fallback values replaced with per-feature training-set medians to prevent test-set leakage.

\section{Hallucination Span Statistics}
\label{app:span_stats}

Across the full RAGTruth dataset (17,790 examples, 2.3M tokens), we identify 13,277 contiguous hallucination spans.
The median span length is 5 tokens (mean 9.4, std 14.8).
Only 10.5\% of spans are single-token; 40.6\% span 2--5 tokens, 39.8\% span 6--20 tokens, and 9.2\% exceed 20 tokens.
Label transition probabilities: $P(H_t \mid H_{t-1}) = 0.902$, $P(H_t \mid F_{t-1}) = 0.006$, giving a persistence ratio of 150:1.
The expected sojourn time in state $H$ is $1/(1 - 0.902) = 10.2$ tokens, consistent with the observed mean span.

\section{Per-Task Performance}
\label{app:per_task}

With BiGRU + all signals (10-seed means): QA $0.887 \pm 0.009$, Data2txt $0.828 \pm 0.015$, Summary $0.760 \pm 0.011$.
All pairwise differences are significant (Wilcoxon $p = 0.002$).
The temporal advantage is largest for Data2txt, where multi-token spans (fabricated statistics, entity substitutions) are most common.

\section{Sequence Length Effect}
\label{app:length}

\begin{table}[H]
\centering
\small
\begin{tabular}{lrccr}
\toprule
\textbf{Quartile} & \textbf{Mean Len.} & \textbf{LogReg} & \textbf{BiGRU} & \textbf{$\Delta$} \\
\midrule
Q1 (short) &  62 & 0.781 & 0.863 & +0.081 \\
Q2         & 103 & 0.726 & 0.814 & +0.087 \\
Q3         & 135 & 0.708 & 0.824 & +0.116 \\
Q4 (long)  & 203 & 0.730 & 0.851 & +0.122 \\
\bottomrule
\end{tabular}
\caption{Token AUC by sequence length quartile. The BiGRU--LogReg gap grows from +8.1 (Q1) to +12.2 (Q4) as longer sequences provide more temporal context.}
\label{tab:length}
\end{table}

\section{Probability Evolution Through Spans}
\label{app:span_evolution}

For each of the 1,305 multi-token hallucination spans in the test set, we extract model predictions from 5 tokens before onset through 15 tokens into the span.
BiGRU shows the highest probability at every position and a rising trajectory (0.50 pre-onset to 0.62 at position 15).
ForwardGRU exhibits the steepest rise (0.42 to 0.57), consistent with forward evidence accumulation.
BackwardGRU starts higher pre-onset (0.49) because it has read the span from the right.

\section{Feature Dynamics at Onset}
\label{app:feature_onset}

\begin{figure}[H]
\centering
\includegraphics[width=\linewidth]{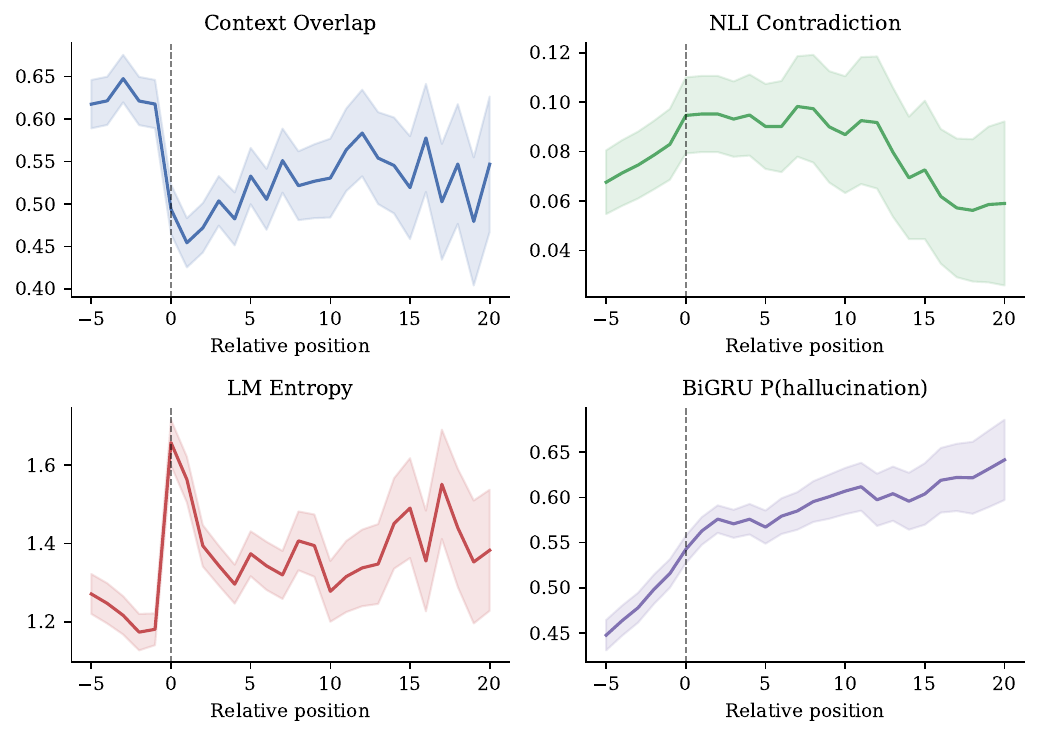}
\caption{Feature dynamics aligned to hallucination span onset (1,136 spans). LM entropy spikes at onset; context overlap drops; NLI contradiction rises gradually; BiGRU probability accumulates monotonically.}
\label{fig:feature_onset}
\end{figure}

Aligned across 1,136 spans (length $\geq 3$): LM entropy shows a +0.48 spike at onset (from 1.18 to 1.66). Context overlap drops by $-0.12$. NLI contradiction rises gradually (+0.012 at onset), consistent with sentence-level granularity. BiGRU probability rises monotonically from 0.45 (pre-onset) to 0.65 (position 20). See Figure~\ref{fig:feature_onset} for the full visualization.

\section{Natural Counterfactual Details}
\label{app:counterfactual}

Same-model pairs (7,492 tokens from 69 queries): NLI contradiction +0.052 ($p < 10^{-67}$), 26 times stronger than cross-model (+0.002). LM entropy shrinks from +0.295 (cross-model) to +0.090 (same-model), indicating that cross-model LM differences partly reflect generator style. Five reconstructed counterfactuals confirm LM entropy spikes of up to +4.5 nats at the exact divergence point.

\section{Onset Detection}
\label{app:onset}

\begin{table}[H]
\centering
\small
\begin{tabular}{@{}lcccc@{}}
\toprule
\textbf{Model} & \textbf{Prec@3} & \textbf{Rec@3} & \textbf{F1@3} & \textbf{Early} \\
\midrule
LogReg      & .100 & .517 & .155 & \textbf{.423} \\
1D-CNN      & .147 & \textbf{.520} & \textbf{.208} & .412 \\
BiGRU       & \textbf{.171} & .298 & .196 & .279 \\
\bottomrule
\end{tabular}
\caption{Onset detection metrics (all signals, tolerance $k=3$). 1D-CNN achieves the best F1 due to high recall; BiGRU has the best precision.}
\label{tab:onset}
\end{table}

\section{Hyperparameter Sweep}
\label{app:sweep}

We sweep hidden dimension $h \in \{32, 64, 128, 256\}$ and number of layers $\in \{1, 2, 3\}$ for BiGRU with all signals (10 seeds each). Training uses AdamW with lr=$10^{-3}$, following \citet{reimers2017optimal}.

\begin{table}[H]
\centering
\small
\caption{BiGRU hyperparameter sweep (mean $\pm$ std AUC, 10 seeds). Performance plateaus at $h{=}64$ and degrades for $h \geq 128$.}
\label{tab:sweep}
\begin{tabular}{@{}rrcr@{}}
\toprule
$h$ & Layers & AUC & Params \\
\midrule
32 & 1 & $.837 \pm .005$ & 15K \\
32 & 2 & $.842 \pm .004$ & 34K \\
32 & 3 & $.840 \pm .007$ & 53K \\
64 & 1 & $.841 \pm .004$ & 46K \\
\textbf{64} & \textbf{2} & $\mathbf{.839 \pm .005}$ & \textbf{121K} \\
64 & 3 & $.843 \pm .005$ & 195K \\
128 & 1 & $.836 \pm .006$ & 158K \\
128 & 2 & $.832 \pm .009$ & 455K \\
128 & 3 & $.830 \pm .011$ & 751K \\
256 & 1 & $.831 \pm .014$ & 579K \\
256 & 2 & $.816 \pm .006$ & 1.8M \\
256 & 3 & $.826 \pm .011$ & 2.9M \\
\bottomrule
\end{tabular}
\end{table}

Performance is stable across $h \in \{32, 64\}$ and layers $\in \{1, 2, 3\}$, with all configurations achieving $0.837$--$0.843$ AUC. For $h \geq 128$, AUC degrades and variance increases, consistent with the 33-dimensional input being too low-dimensional to support large hidden states. We select $h{=}64$, 2 layers (121K parameters) as the default configuration because it matches the MLP baseline parameter count (112K). This allows a controlled capacity comparison.

\section{Robustness Across Seeds}
\label{app:robustness_fig}

\begin{figure}[H]
\centering
\includegraphics[width=\linewidth]{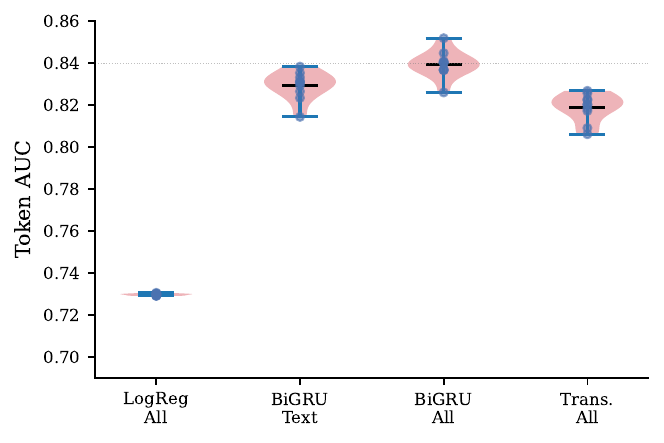}
\caption{AUC distribution across 10 seeds for key architectures. Temporal models (BiGRU, BiLSTM) consistently outperform non-temporal baselines (LogReg, MLP).}
\label{fig:robustness}
\end{figure}

\section{Training Recipe Ablation}
\label{app:recipe}

We test six training recipes on BiGRU $h{=}64$ L=2 (10 seeds each): BCE vs.\ Focal Loss ($\gamma{=}2$), and constant vs.\ cosine (5\% warmup) vs.\ ReduceLROnPlateau (patience=3, factor=0.5) learning rate schedules.

\begin{table}[H]
\centering
\small
\caption{Training recipe ablation on BiGRU; the first row (BCE + constant) is the baseline. No recipe significantly outperforms it, suggesting the AUC ceiling is determined by the features rather than the optimization.}
\label{tab:recipe}
\begin{tabular}{@{}lcccc@{}}
\toprule
\textbf{Recipe} & \textbf{AUC} & \textbf{F1} & \textbf{P} & \textbf{R} \\
\midrule
BCE + constant   & $.839 \pm .005$ & .254 & .158 & .654 \\
BCE + plateau    & $.837 \pm .006$ & .249 & .154 & .661 \\
BCE + cosine     & $.840 \pm .006$ & .248 & .154 & .661 \\
Focal + constant & $.840 \pm .006$ & .248 & .152 & .670 \\
Focal + plateau  & $.839 \pm .007$ & .249 & .153 & .663 \\
Focal + cosine   & $.839 \pm .007$ & .242 & .148 & .675 \\
\bottomrule
\end{tabular}
\end{table}

Focal Loss increases recall (+0.02 over BCE) at the cost of precision, consistent with its design: downweighting easy negatives shifts the decision boundary toward the positive class. No recipe breaks through the ${\sim}0.840$ AUC ceiling, indicating that performance is limited by the 33-dimensional feature representation rather than by the optimization procedure.

The same pattern holds for the Transformer encoder: all six recipes yield AUC in $[0.796, 0.805]$ (Table~\ref{tab:recipe_transformer}), confirming that the ${\sim}0.80$ Transformer ceiling is likewise architectural, not optimization-driven.

\begin{table}[H]
\centering
\small
\caption{Training recipe ablation on Transformer Pre-LN (10 seeds each).}
\label{tab:recipe_transformer}
\begin{tabular}{@{}lcc@{}}
\toprule
\textbf{Recipe} & \textbf{AUC} & \textbf{F1} \\
\midrule
BCE + constant   & $.805 \pm .008$ & .204 \\
BCE + plateau    & $.803 \pm .011$ & .203 \\
BCE + cosine     & $.796 \pm .008$ & .198 \\
Focal + constant & $.799 \pm .008$ & .209 \\
Focal + plateau  & $.801 \pm .010$ & .206 \\
Focal + cosine   & $.797 \pm .009$ & .201 \\
\bottomrule
\end{tabular}
\end{table}

\section{Full Ablation Heatmap}
\label{app:heatmap}

\begin{figure}[H]
\centering
\includegraphics[width=\linewidth]{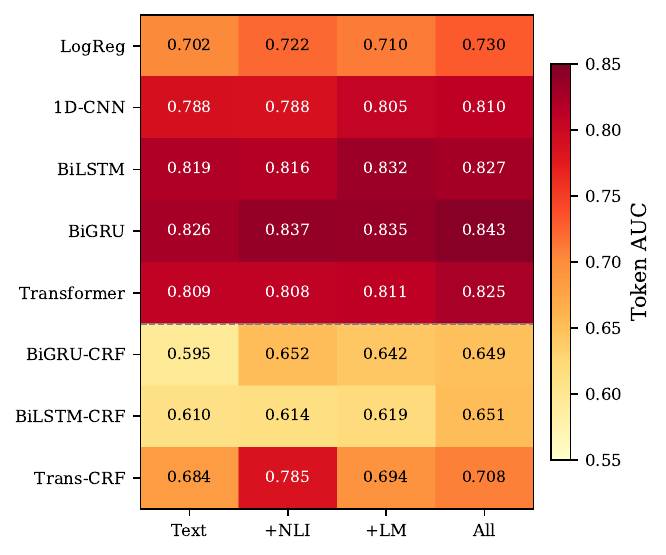}
\caption{Token AUC across all architecture--signal combinations (single seed). Non-CRF temporal models (BiGRU, BiLSTM) consistently outperform baselines across all signal configurations.}
\label{fig:heatmap}
\end{figure}

\section{Cross-Model Transfer Details}
\label{app:cross_model}

\begin{figure}[H]
\centering
\includegraphics[width=0.85\linewidth]{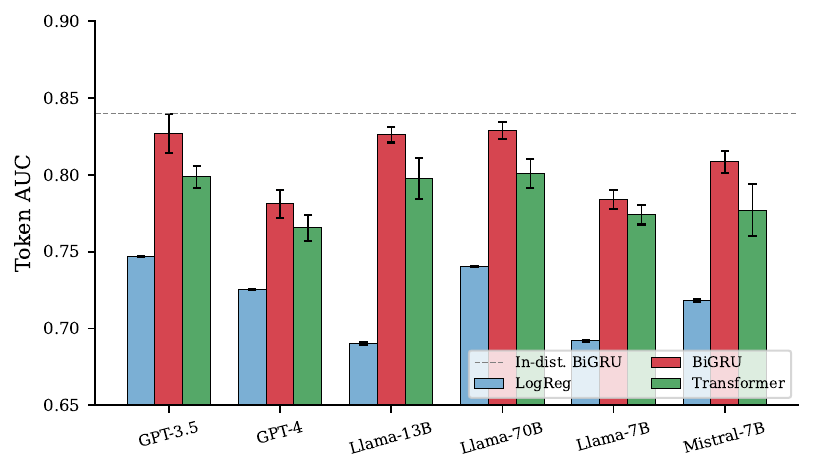}
\caption{Leave-one-out cross-model generalization (all signals, 10 seeds). GPT-4 is the hardest transfer target; LLaMA-2-70B is the easiest.}
\label{fig:cross_model}
\end{figure}

\section{Lookback Lens Full Results}
\label{app:lookback}

\begin{table}[H]
\centering
\small
\begin{tabular}{llccc}
\toprule
\textbf{Features} & \textbf{Model} & \textbf{AUC} & \textbf{F1} & \textbf{AP} \\
\midrule
\multicolumn{5}{l}{\emph{TinyLlama-1.1B proxy}} \\
Lookback       & LogReg & $.699$ & $.194$ & $.173$ \\
Lookback       & BiGRU  & $.822$ & $.355$ & $.344$ \\
Combined       & BiGRU  & $.853$ & $.355$ & $.401$ \\
\midrule
\multicolumn{5}{l}{\emph{Qwen3-14B proxy}} \\
Lookback       & LogReg & $.726$ & $.204$ & $.187$ \\
Lookback       & BiGRU  & $.838$ & $.384$ & $.368$ \\
Combined       & BiGRU  & $\mathbf{.866}$ & $\mathbf{.391}$ & $\mathbf{.425}$ \\
\midrule
Ours only      & BiGRU  & $.819$ & $.297$ & $.298$ \\
\bottomrule
\end{tabular}
\caption{Full Lookback Lens comparison (open-LLM subset, 5 seeds).}
\label{tab:lookback_full}
\end{table}

\section{Signal Contribution Analysis}
\label{app:signal}

\begin{figure}[H]
\centering
\includegraphics[width=\linewidth]{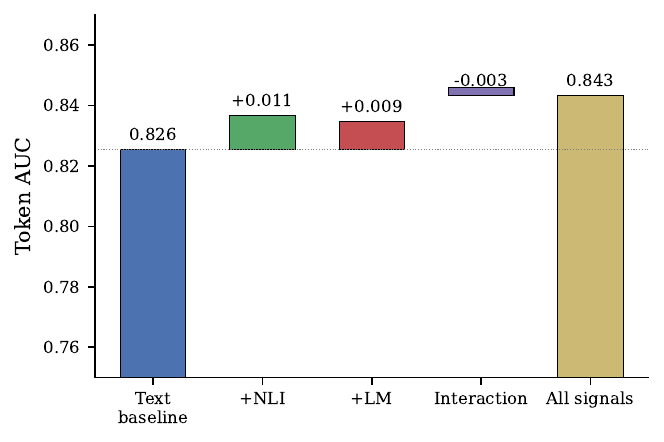}
\caption{Waterfall decomposition of signal contributions to BiGRU's AUC. NLI and LM features provide complementary gains beyond text features alone.}
\label{fig:signal_contribution}
\end{figure}

\section{Precision-Recall Curves}
\label{app:pr}

\begin{figure}[H]
\centering
\includegraphics[width=0.85\linewidth]{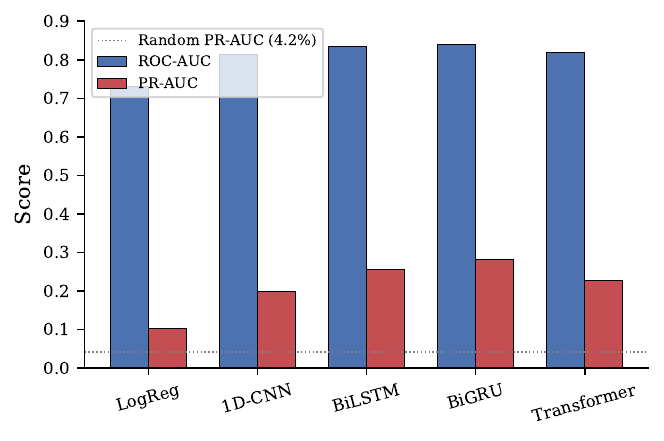}
\caption{Precision-recall curves for all non-CRF architectures (all signals, seed 42). The low overall precision reflects RAGTruth's 5.6\% hallucination rate.}
\label{fig:pr_curves}
\end{figure}

\section{Per-Task Comparison}
\label{app:per_task_fig}

\begin{figure}[H]
\centering
\includegraphics[width=\linewidth]{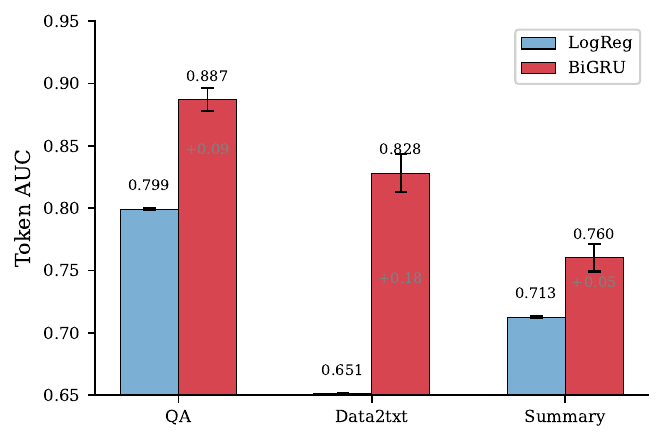}
\caption{Per-task AUC comparison (all signals). The temporal advantage of BiGRU over LogReg is largest for Data2txt.}
\label{fig:per_task}
\end{figure}

\section{Sequence Length Analysis}
\label{app:length_fig}

\begin{figure}[H]
\centering
\includegraphics[width=0.85\linewidth]{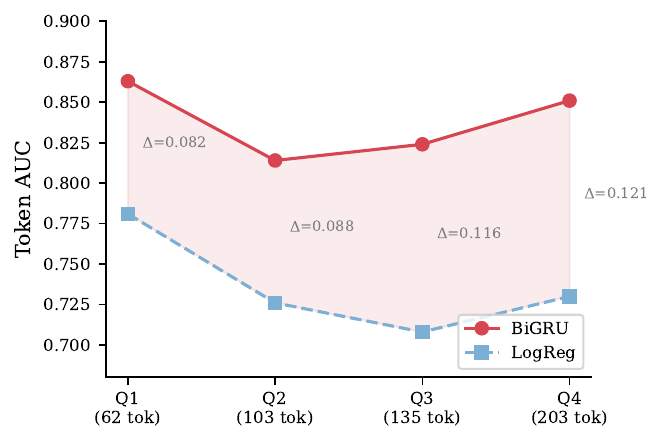}
\caption{Token AUC by sequence length quartile. The BiGRU--LogReg gap grows from +8.1 (Q1) to +12.2 (Q4).}
\label{fig:length}
\end{figure}

\section{CRF Calibration Details}
\label{app:crf_fig}

\begin{figure}[H]
\centering
\includegraphics[width=0.85\linewidth]{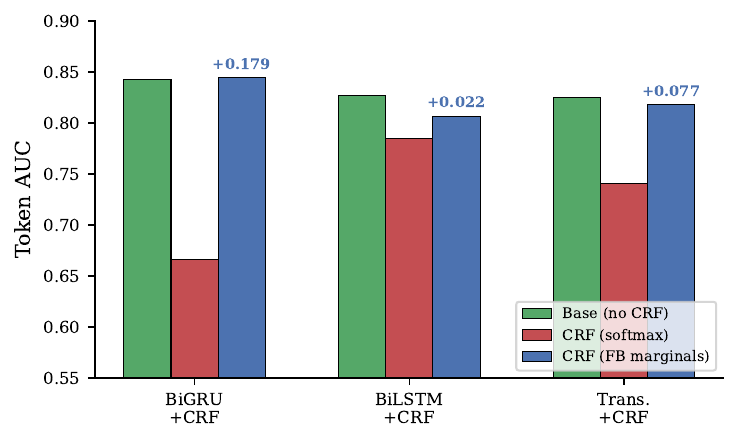}
\caption{CRF scoring methods compared. Forward-backward marginals recover up to +17.9 AUC points over softmax scoring.}
\label{fig:crf}
\end{figure}

\section{Counterfactual Feature Comparison}
\label{app:counterfactual_fig}

\begin{figure}[H]
\centering
\includegraphics[width=\linewidth]{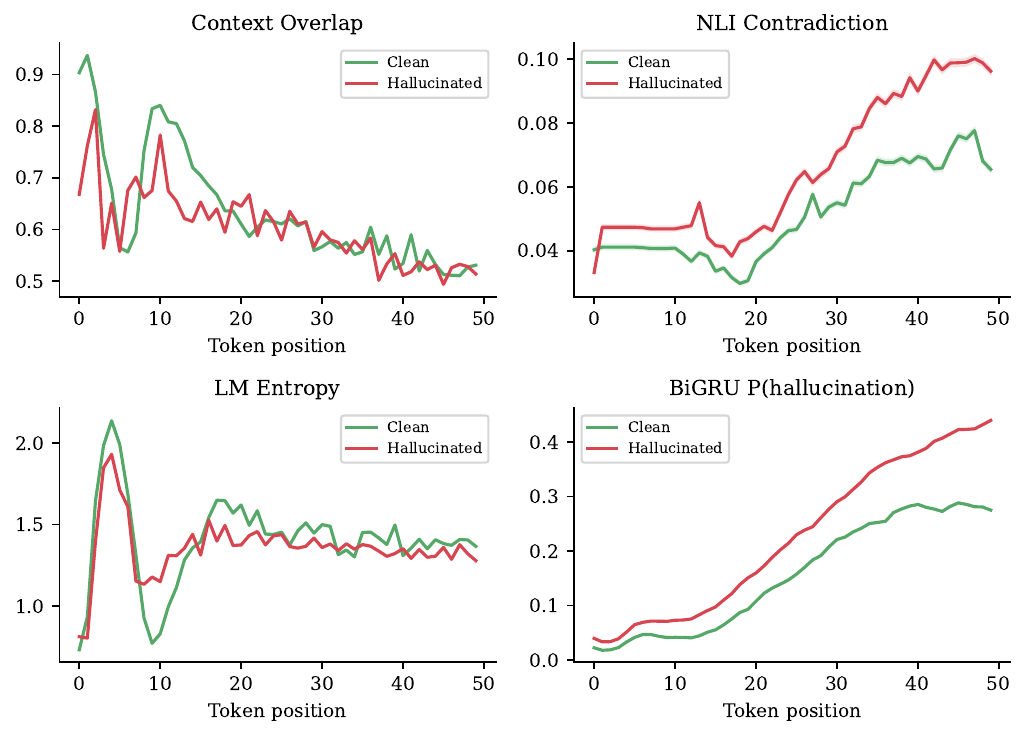}
\caption{Natural counterfactual: same query, different LLMs, one hallucinating. NLI contradiction separates the two; LM entropy does not; BiGRU integrates both.}
\label{fig:counterfactual}
\end{figure}

\section{Transfer Comparison}
\label{app:transfer_fig}

\begin{figure}[H]
\centering
\includegraphics[width=0.85\linewidth]{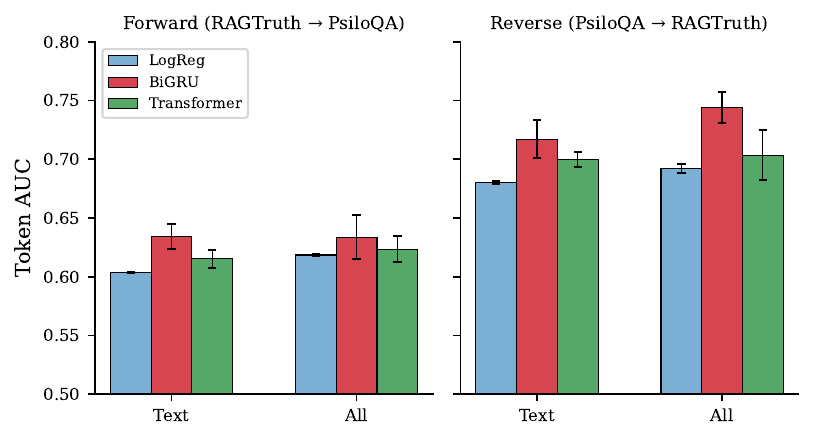}
\caption{Bidirectional dataset transfer. Reverse transfer (PsiloQA to RAGTruth) outperforms forward despite fewer training examples.}
\label{fig:transfer}
\end{figure}

\section{Emerging Architecture Comparison}
\label{app:arch_comparison}

We evaluate seven architectures, each with its literature-recommended training recipe (Table~\ref{tab:arch_cosine}), and additionally with ReduceLROnPlateau + early stopping (patience 10, max 50 epochs; Table~\ref{tab:arch_plateau}).

\begin{table}[H]
\centering
\small
\caption{Architecture comparison with cosine schedule (literature recipes, 5--10 seeds). BiGRU, Mamba, and BiGRU+Attention converge to $\sim$0.84.}
\label{tab:arch_cosine}
\begin{tabular}{@{}lcccc@{}}
\toprule
\textbf{Architecture} & \textbf{AUC} & \textbf{F1} & \textbf{Params} & \textbf{Seeds} \\
\midrule
\textbf{BiGRU} & $\mathbf{.840 \pm .007}$ & .257 & 121K & 10 \\
BiGRU+Attention & $.840 \pm .005$ & .251 & 113K & 10 \\
Mamba & $.838 \pm .006$ & .244 & 203K & 9 \\
BiLSTM & $.835 \pm .008$ & .242 & 121K & 10 \\
BixLSTM & $.827 \pm .006$ & .233 & 125K & 10 \\
DilatedCNN & $.826 \pm .006$ & .244 & 61K & 10 \\
Transformer & $.796 \pm .008$ & .204 & 106K & 10 \\
\bottomrule
\end{tabular}
\end{table}

\begin{table}[H]
\centering
\small
\caption{Same architectures with ReduceLROnPlateau + early stopping (max 50 epochs, patience 10, 5 seeds). Plateau scheduling raises the ceiling from 0.840 to 0.845 and eliminates most architecture differences.}
\label{tab:arch_plateau}
\begin{tabular}{@{}lcccc@{}}
\toprule
\textbf{Architecture} & \textbf{AUC} & \textbf{$\Delta$} & \textbf{Best Ep} & \textbf{Params} \\
\midrule
BiLSTM & $\mathbf{.845 \pm .003}$ & +.010 & 17 & 121K \\
Mamba & $.844 \pm .005$ & +.006 & 20 & 203K \\
BiGRU+Attn & $.844 \pm .002$ & +.004 & 12 & 113K \\
BiGRU & $.843 \pm .003$ & +.003 & 8 & 121K \\
BixLSTM & $.843 \pm .004$ & +.016 & 19 & 125K \\
DilatedCNN & $.833 \pm .002$ & +.007 & 13 & 61K \\
Transformer & $.827 \pm .002$ & +.031 & 46 & 106K \\
\bottomrule
\end{tabular}
\end{table}

With plateau scheduling, the top five architectures (BiLSTM, Mamba, BiGRU+Attention, BiGRU, BixLSTM) all achieve 0.843--0.845 AUC, confirming that the ceiling is determined by the 33-dimensional feature representation rather than by the model architecture. The largest gains are for architectures that were undertrained with fixed-epoch cosine: Transformer (+0.031), BixLSTM (+0.016), BiLSTM (+0.010). BiGRU converges fastest (best epoch 8) and remains the best choice for deployment efficiency.

\begin{figure}[H]
\centering
\includegraphics[width=\linewidth]{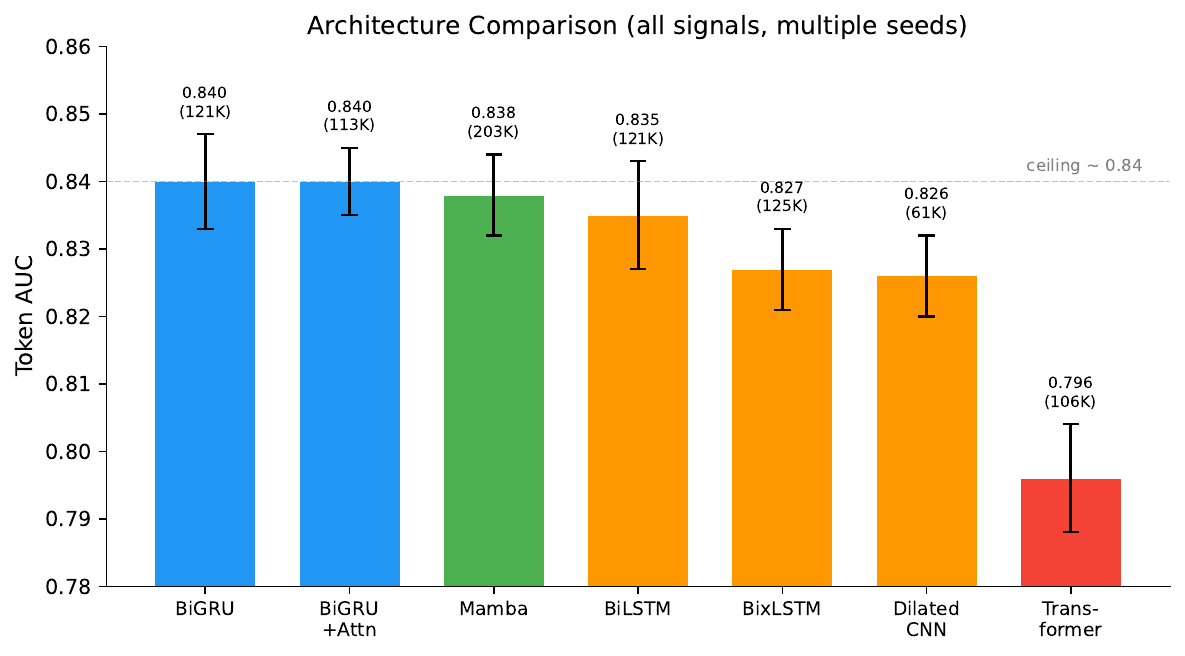}
\caption{Architecture comparison (cosine schedule). All temporal architectures with full-sequence scope converge to $\sim$0.84 AUC. With plateau scheduling (Table~\ref{tab:arch_plateau}), the ceiling rises to 0.845.}
\label{fig:arch_comparison}
\end{figure}

\section{Probing Comparison}
\label{app:probing}

A reviewer may ask whether probing the generating model's internal states already captures the temporal information our method exploits.
We test this directly by extracting hidden states from Mistral-7B-Instruct (one of the RAGTruth generators) and comparing probing approaches against our method on the Mistral-generated subset.

\begin{table}[H]
\centering
\caption{Probing comparison: hidden states from the generating model vs.\ our 33-dim black-box features. Each generator is evaluated only on its own outputs. Hidden states are extracted from 4 evenly spaced layers (dim=16,384).}
\label{tab:probing}
\small
\begin{tabular}{@{}lcccc@{}}
\toprule
 & \multicolumn{2}{c}{\textbf{LLaMA-2-7B}} & \multicolumn{2}{c}{\textbf{Mistral-7B}} \\
\cmidrule(lr){2-3} \cmidrule(lr){4-5}
\textbf{Method} & AUC & F1 & AUC & F1 \\
\midrule
Linear probe on HS & .543 & .118 & .574 & .142 \\
BiGRU on HS & .637 & .195 & .631 & .201 \\
\midrule
\textbf{BiGRU on 33-dim (ours)} & $\mathbf{.770}$ & $\mathbf{.245}$ & $\mathbf{.833}$ & $\mathbf{.355}$ \\
BiGRU on 33-dim + HS & .748 & .215 & .793 & .282 \\
\bottomrule
\end{tabular}
\end{table}

Three findings are consistent across both generators.
First, probing hidden states yields 0.543--0.574 AUC with a linear probe and 0.631--0.637 with BiGRU---far below our black-box method (0.770--0.833).
The generator's internal representations do not reliably encode hallucination status: the model does not ``know'' it is hallucinating, which is precisely why it hallucinates.
Second, adding temporal modeling to probing (BiGRU vs.\ linear probe) improves AUC by +0.06--0.09, showing that temporal structure exists in hidden state sequences too, but the signal is too weak to be useful on its own.
Third, combining our features with hidden states (0.748--0.793) is \emph{worse} than our features alone (0.770--0.833), indicating that the noisy hidden-state signal degrades performance when concatenated with our engineered features.

These results directly address the concern that ``summarization has already happened'' in internal representations: while the representations encode sequential context via self-attention, they do not encode the \emph{hallucination detection signal} that our external features capture.

\paragraph{Does layer selection or all-layer fusion rescue probing?}
A natural objection~\citep{wang2026fepoid} is that the hallucination signal is concentrated in a specific intermediate layer, so probing four evenly spaced layers may understate what hidden states offer. We test this thoroughly on LLaMA-2-7B-chat, at the token level, on its own outputs (a smaller $n{=}450$ test subset, since extracting and storing all 32 layers' hidden states is costly). (i)~Probing \emph{all 32 layers} separately, the best single layer (chosen on the test set, the most generous selection) reaches only $0.527$ AUC with a linear probe and $0.590$ with a BiGRU. (ii)~A \emph{learned softmax fusion of all 32 layers} feeding a deep MLP reaches $0.518$ and feeding a BiGRU $0.562$; the learned layer weights are near-uniform (max $0.032 \approx 1/32$), so there is no informative layer or combination to find, and fusing all layers does not beat the best single one. Against $0.747$ for our 33-dim features on the same subset, neither layer selection nor all-layer fusion rescues \emph{token-level} probing: the response-level signal these methods exploit (AUROC ${\sim}0.87$ in \citealt{wang2026fepoid}) is not recovered per token by any probe we tried (linear, best-layer, or learned fusion, with an MLP or a BiGRU).

\section{Per-Task Architecture Comparison}
\label{app:per_task_archs}

\begin{table}[H]
\centering
\caption{Per-task AUC for five architectures (3 seeds, plateau+ES). QA is easiest across all architectures; Data2txt shows the largest temporal advantage.}
\label{tab:per_task_archs}
\small
\begin{tabular}{@{}lccc@{}}
\toprule
\textbf{Architecture} & \textbf{QA} & \textbf{Data2txt} & \textbf{Summary} \\
\midrule
Mamba & $.890{\pm}.007$ & $.827{\pm}.009$ & $.766{\pm}.005$ \\
BiGRU & $.885{\pm}.002$ & $\mathbf{.833{\pm}.006}$ & $.758{\pm}.017$ \\
BiGRU+Attn & $.885{\pm}.004$ & $.830{\pm}.008$ & $.770{\pm}.010$ \\
BixLSTM & $.884{\pm}.003$ & $.815{\pm}.017$ & $\mathbf{.771{\pm}.013}$ \\
DilatedCNN & $.883{\pm}.005$ & $.795{\pm}.009$ & $.760{\pm}.008$ \\
\bottomrule
\end{tabular}
\end{table}

All architectures achieve $\sim$0.88 AUC on QA, where factual errors produce sharp signals.
The temporal advantage is most pronounced on Data2txt, where BiGRU (0.833) outperforms DilatedCNN (0.795) by 3.8 points---multi-token hallucination spans (fabricated statistics, entity substitutions) benefit most from full-sequence temporal modeling.
Summary is hardest ($\sim$0.76), consistent with the difficulty of detecting subtle omissions and distortions at the token level.

\end{document}